\documentclass[sigconf]{acmart}
\usepackage{booktabs}
\usepackage[linesnumbered,ruled,vlined,algo2e]{algorithm2e}
\hypersetup{
  hidelinks,
  unicode=true,
  bookmarksnumbered=true
}
\usepackage{url}
\usepackage{fontawesome5}
\usepackage[ruled,vlined]{algorithm2e}
\usepackage{tabularx}
\usepackage{multirow}    
\usepackage{tabularx}    
\usepackage{array}       
\usepackage{graphicx}    
\usepackage{float}
\usepackage{capt-of}
\usepackage{afterpage}      
\usepackage{placeins}
\usepackage{booktabs,tabularx,array}
\newcolumntype{Z}{>{\raggedright\arraybackslash}X}
\newcolumntype{Y}{>{\raggedright\arraybackslash}X} 
\newcolumntype{R}{>{\raggedleft\arraybackslash}p{1.25cm}}
\newcommand{\sourcelink}[1]{\href{#1}{URL}}
\usepackage{placeins}
\usepackage{dblfloatfix} 
\usepackage{cuted}

\usepackage{xcolor}
\usepackage{tabularx}
\usepackage{booktabs}
\usepackage{colortbl}

\definecolor{groupblue}{RGB}{198,240,240}

\definecolor{stockred}{HTML}{DD4B39}
\definecolor{stockgreen}{HTML}{00A65A}

\newcommand{\pos}[1]{{\color{stockred}\scriptsize(+#1)}}
\newcommand{\dneg}[1]{{\color{stockgreen}\scriptsize(\textminus#1)}}
\newcommand{\posbold}[1]{{\color{stockred}\scriptsize\textbf{(+#1)}}}

\newcolumntype{Y}{>{\centering\arraybackslash}X}

\newcommand{\grouprow}[1]{%
  \noalign{\vskip -0.8ex}
  \rowcolor{groupblue}%
  \multicolumn{7}{l}{\rule{0pt}{2.3ex}\textit{#1}}\\[-0.3em]
}

\newcommand{\grouprowopen}[1]{%
  \noalign{\vskip -0.8ex}%
  \rowcolor{groupblue}%
  \multicolumn{5}{l}{\rule{0pt}{2.3ex}\textit{#1}}\\[-0.3em]
}

\usepackage{capt-of}     
\usepackage{enumitem}
\usepackage{booktabs}   
\usepackage{tabularx}  
\usepackage{xcolor}    
\usepackage{array}     

\definecolor{oursblue}{HTML}{2E86C1}
\AtBeginDocument{%
  }

\setcopyright{acmlicensed}
\copyrightyear{2018}
\acmYear{2018}
\acmDOI{XXXXXXX.XXXXXXX}
\acmConference[Conference acronym 'XX]{Make sure to enter the correct
  conference title from your rights confirmation email}{June 03--05,
  2018}{Woodstock, NY}
\acmISBN{978-1-4503-XXXX-X/2018/06}
\begin{document}

\title{MM-NeuroOnco: A Multimodal Benchmark and Instruction Dataset for MRI-Based Brain Tumor Diagnosis}

\author{Feng Guo}
\authornote{These authors contributed equally to this work.}
\email{guofeng@gdiist.cn}
\affiliation{%
  \institution{Guangdong Institute of Intelligence Science and Technology}
  \city{Hengqin}
  \country{China}
}

\author{Jiaxiang Liu}
\authornotemark[1]
\email{liujiaxiang@gdiist.cn}
\affiliation{%
  \institution{Guangdong Institute of Intelligence Science and Technology}
  \city{Hengqin}
  \country{China}
}

\author{Yang Li}
\email{liyang@gdiist.cn}
\affiliation{%
  \institution{Guangdong Institute of Intelligence Science and Technology}
  \city{Hengqin}
  \country{China}
}

\author{Qianqian Shi}
\email{qqshi@mail.tsinghua.edu.cn}
\affiliation{%
  \institution{Center for Brain-Inspired Computing Research (CBICR), Department of Precision Instrument, Tsinghua University}
  \city{Beijing}
  \country{China}
}

\author{Mingkun Xu}
\authornote{Corresponding author.}
\email{xumingkun@gdiist.cn}
\affiliation{%
  \institution{Guangdong Institute of Intelligence Science and Technology}
  \city{Hengqin}
  \country{China}
}

\renewcommand{\shortauthors}{Guo, Liu, Yang, and Xu}

\begin{abstract}
Accurate brain tumor diagnosis requires models to not only detect lesions but also generate clinically interpretable reasoning grounded in imaging manifestations, yet existing public datasets remain limited in annotation richness and diagnostic semantics. To bridge this gap, we introduce \textbf{MM-NeuroOnco}, a large-scale multimodal benchmark and instruction-tuning dataset for brain tumor MRI understanding, consisting of 24,726 MRI slices from 20 data sources paired with approximately 200,000 semantically enriched multimodal instructions spanning diverse tumor subtypes and imaging modalities. To mitigate the scarcity and high cost of diagnostic semantic annotations, we develop a multi-model collaborative pipeline for automated medical information completion and quality control, enabling the generation of diagnosis-related semantics beyond mask-only annotations. Building upon this dataset, we further construct \textbf{MM-NeuroOnco-Bench}, a manually annotated evaluation benchmark with a rejection-aware setting to reduce biases inherent in closed-ended question formats. Evaluation across ten representative models shows that even the strongest baseline, Gemini~3~Flash, achieves only 41.88\% accuracy on diagnosis-related questions, highlighting the substantial challenges of multimodal brain tumor diagnostic understanding. Leveraging MM-NeuroOnco, we further propose \textbf{NeuroOnco-GPT}, which achieves a 27\% absolute accuracy improvement on diagnostic questions following fine-tuning. This result demonstrates the effectiveness of our dataset and benchmark in advancing clinically grounded multimodal diagnostic reasoning. Code and dataset are publicly available at:   \url{https://github.com/gfnnnb/MM-NeuroOnco}.
\end{abstract}



\maketitle

\section{Introduction}

Brain tumors are a highly lethal class of diseases arising within the central nervous system~\cite{bray2024global,price2024cbtrus}. 
Clinical practice indicates that reliable differential diagnosis cannot be achieved through lesion localization alone, but instead requires holistic reasoning grounded in multi-dimensional imaging semantics~\cite{csutak2020differentiating}. 
In recent years, deep learning has achieved remarkable progress in brain tumor segmentation tasks, with paradigms such as the BraTS challenge series continuously advancing pixel-level lesion modeling capabilities~\cite{menze2014multimodal}. 
However, these segmentation-centric approaches primarily emphasize spatial boundary delineation, while largely overlooking the medical semantic modeling and diagnostic reasoning that are essential for clinical decision-making~\cite{setyawan2024beyond}. 
High segmentation accuracy does not necessarily translate into correct understanding of tumor invasiveness, cross-modal discrepancies, or pathological implications, leaving existing models with notable limitations in real-world clinical scenarios.

\begin{figure*}[t]
  \centering
  \includegraphics[width=0.9\textwidth]{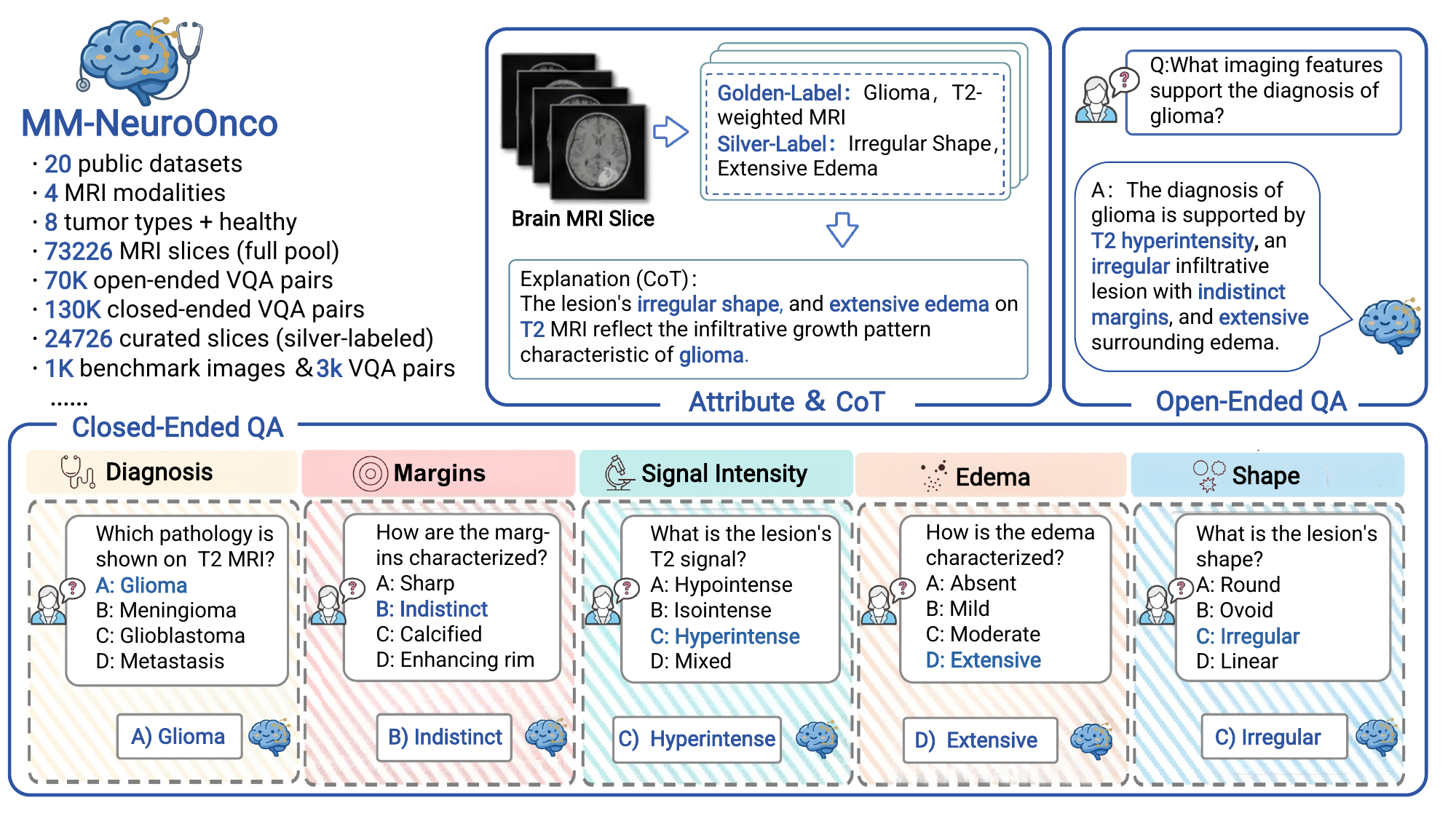}
  \caption{Overview of MM-NeuroOnco. The benchmark curates 24,726 slices from 20 diverse datasets. Standard diagnostic labels are augmented with fine-grained semantic attributes to construct explicit Chain-of-Thought (CoT) reasoning. This structure enables a multi-grained evaluation paradigm, covering both holistic open-ended diagnosis and targeted closed-ended Question Answering across multiple clinical categories.}
  \Description{Overview of MM-NeuroOnco, showing dataset curation across 20 sources, semantic attribute augmentation, and CoT-style reasoning for multi-grained evaluation (open-ended diagnosis and closed-ended QA across clinical categories).}
  \label{fig:figure1}
\end{figure*}

In routine clinical workflows, radiologists typically base brain tumor diagnosis on a small set of highly informative two-dimensional slices, interpreted through cross-modal comparison across MRI sequences, rather than exhaustive voxel-wise analysis of full three-dimensional volumes. 
This slice-centered diagnostic paradigm is particularly aligned with auxiliary diagnosis and clinical settings requiring rapid diagnosis, and more closely reflects how human clinicians perform diagnostic reasoning in practice. 
By contrast, directly operating on full 3D volumes, while offering richer spatial context, often incurs substantial computational overhead and inference latency, limiting its practicality in efficient clinical deployment~\cite{dorfner2025review}.

To address these limitations, as illustrated in Figure~\ref{fig:figure1}, we construct a multimodal instruction-tuning dataset and benchmark tailored for brain tumor MRI understanding. 
Distinct from prior works that emphasize visual recognition or coarse-grained question answering, our design focuses on high-density diagnostic semantics and structured reasoning supervision. 
By organizing diagnosis-relevant attributes into explicit intermediate reasoning steps, we enable models to follow clinically grounded diagnostic logic, while a more robust evaluation protocol facilitates faithful assessment of model capability boundaries and reliability in high-stakes diagnostic settings. 
The main contributions of this work are threefold:
\begin{itemize}
  \item \textbf{Comprehensive Benchmark \& Dataset:} We construct \textbf{MM-NeuroOnco}, a large-scale multimodal resource comprising 24,726 MRI slices across four modalities, covering eight tumor subtypes and healthy controls. It features approximately 200,000 semantically enriched instruction samples and includes \textbf{MM-NeuroOnco-Bench}, a manually annotated subset for rigorous evaluation.
  \item \textbf{Automated Semantic Completion Pipeline:} We propose a novel data construction framework that leverages multi-model collaboration to generate diagnosis-relevant attributes from sparse annotations. This pipeline effectively alleviates the semantic gap and the prohibitive costs associated with large-scale expert labeling.
  \item \textbf{Specialized Model \& Robust Evaluation:} We develop \textbf{NeuroOnco-GPT}, validating the practical value of our dataset. Furthermore, we introduce a rejection-aware evaluation protocol to mitigate the ``forced-choice'' bias inherent in closed-ended questions, enabling a more faithful assessment of model reliability and capability boundaries in high-stakes diagnostic tasks.
\end{itemize}

\begin{figure*}[t!]
  \centering
  \includegraphics[width=0.99\textwidth]{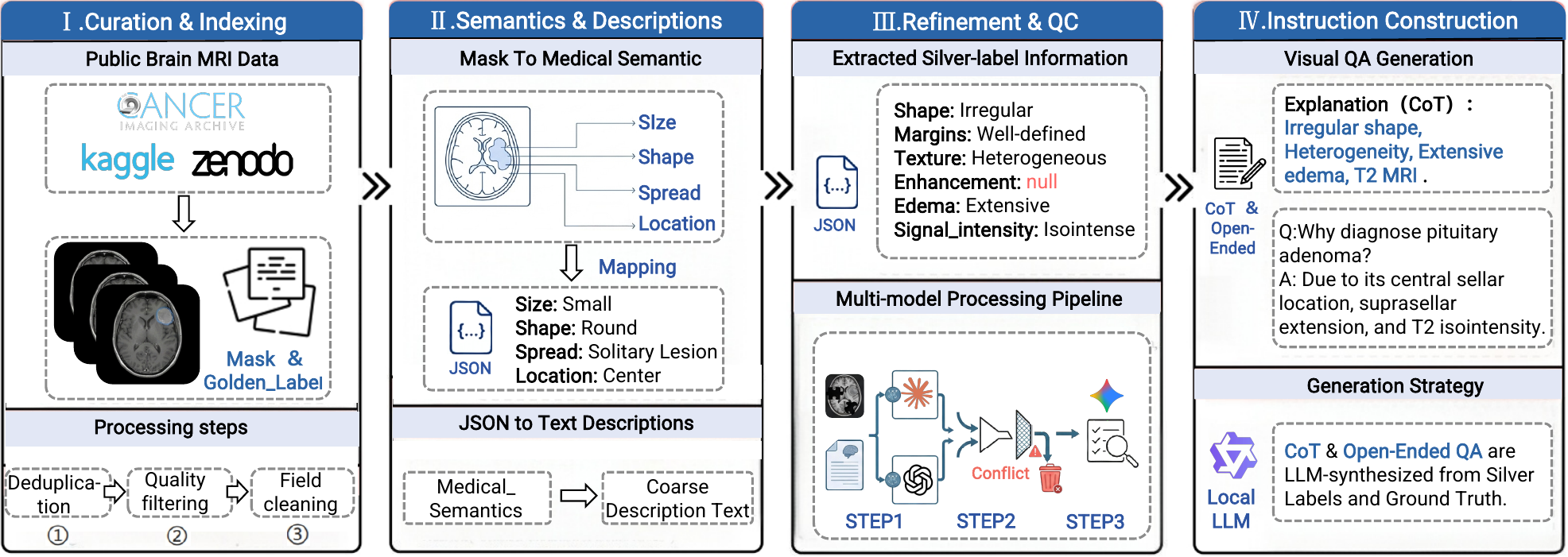}
  \caption{The MM-NeuroOnco dataset curation pipeline, which consists of four sequential steps transforming raw MRI data into semantically enriched multimodal instructions.}
  \Description{The MM-NeuroOnco dataset curation pipeline with four sequential steps that transform raw MRI data into semantically enriched multimodal instructions.}
  \label{fig:figure2}
\end{figure*}

\section{Related Work}
\subsection{Brain Tumor Datasets and Benchmarks}
Brain tumor MRI analysis has long been a focal point of medical imaging research, leading to the establishment of numerous representative public datasets and evaluation benchmarks. 
Early efforts concentrated predominantly on the segmentation and quantitative analysis of tumor regions, with the BraTS (Brain Tumor Segmentation) challenge series emerging as the most influential initiative~\cite{menze2014multimodal,bakas2017advancing,baid2021rsna,adewole2023brain}. 
By consistently releasing multi-year datasets featuring multi-modal MRI scans (e.g., T1, T2, FLAIR, T1ce) alongside fine-grained pixel-level annotations, BraTS has become the gold standard benchmark for brain tumor segmentation research~\cite{menze2014multimodal}. 
Beyond BraTS, {The Cancer Imaging Archive (TCIA) aggregates a diverse array of brain tumor-related MRI collections. 
Spanning various tumor types, imaging protocols, and clinical contexts, these resources have been extensively utilized for tumor segmentation, classification, and multimodal image analysis research~\cite{clark2013cancer}.

Furthermore, the systematic curation and analysis of large-scale medical imaging datasets provide essential context for brain tumor research. 
For instance, Project Imaging-X conducted a systematic review of over 1,000 open medical imaging datasets, performing statistical analyses across dimensions such as modality, task, and anatomical site~\cite{projectimagingx2025}. Overall, these rich resources have provided essential data support for research in this field, significantly propelling its general advancement.

\subsection{Challenges in Medical Evaluation Paradigms}

The choice of evaluation paradigm fundamentally dictates how model capabilities are characterized, yet finding a robust metric for medical reasoning remains an open challenge~\cite{he2023medeval,liu2024medbench}.
Closed-ended evaluation (e.g., multiple-choice) is widely adopted due to its reproducibility and straightforward metrics. 
However, this format is inherently reductionist: it collapses complex clinical reasoning into discrete options. 
Consequently, models often learn to exploit statistical biases in option distributions or language priors rather than grounding their answers in imaging evidence, leading to inflated performance scores that do not reflect true diagnostic capability~\cite{ye2024gmai,yue2024mmmu}.

Conversely, open-ended generation offers a broader response space better suited for assessing reasoning coherence. 
Yet, it suffers from metric instability. 
Determining semantic equivalence between a model's output and a reference answer is notoriously difficult, and standard metrics (such as BLEU or ROUGE) correlate poorly with clinical accuracy~\cite{lau2018dataset,he2020pathvqa}. 
Furthermore, generative models are prone to ``hallucinations''—fabricating convincing but factually incorrect medical details—which undermines trust in automated scoring~\cite{pal2023med,xia2024cares}.

Prior works have attempted to bridge this divide by introducing structured constraints, auxiliary supervision, or explicit reasoning traces~\cite{gai2024medthink,liu2024medcot,wang2025v2t}. 
Despite these efforts, faithfully characterizing diagnostic reasoning while maintaining evaluation reproducibility remains an unresolved bottleneck, particularly in brain tumor scenarios where imaging semantics are highly specialized and subtle.

\section{MM-NeuroOnco Dataset Construction}
The construction of MM-NeuroOnco follows a rigorous four-stage pipeline designed to transform raw, heterogeneous MRI collections into a semantically dense instruction-tuning dataset. The overall workflow is illustrated in Figure~\ref{fig:figure2}.

\subsection{Data Curation \& Standardization}
We aggregated a massive corpus of over 100,000 brain MRI scans from disparate public repositories, including Kaggle, Zenodo, and TCIA. Raw data from these sources suffered from severe heterogenei-ty---ranging from inconsistent modality tagging and chaotic file structures to non-standardized diagnostic labels. To resolve this, we designed a Unified Metadata Schema, converting all samples into a standardized JSON index format. This step effectively eliminates structural discrepancies, establishing a single, consistent access point for diverse data sources.

Following strict deduplication and quality filtering, we established a ``Full Master Index'' containing 73,226 valid samples, of which 19,086 are paired with pixel-level segmentation masks. This Master Index underpins semantic completion and task sampling, forming the basis of both the MM-NeuroOnco training set and the evaluation benchmark.

\begin{figure}[t]
  \centering
  \hspace{0.08\columnwidth}
\includegraphics[width=0.9\columnwidth]{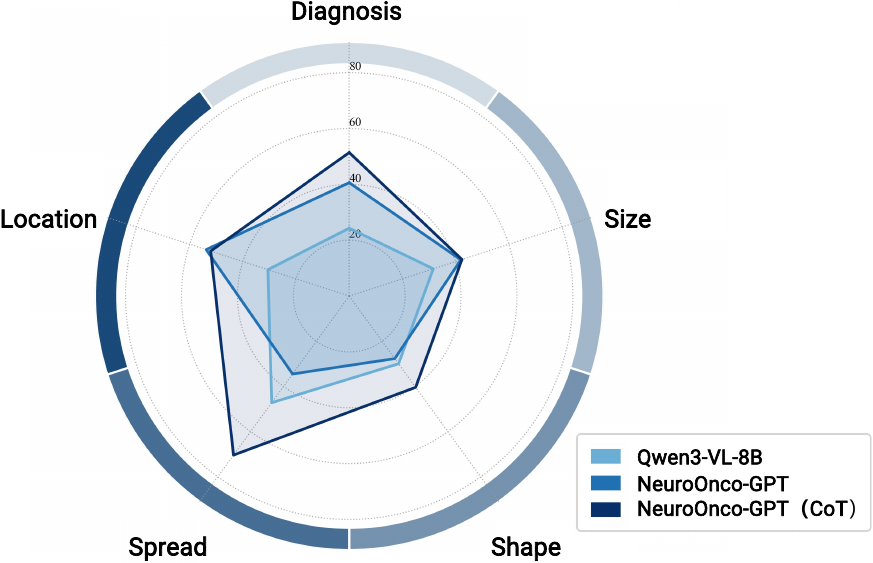}
  \caption{Radar chart comparing the performance of the base model and different fine-tuning strategies on MM-NeuroOnco-Bench.}
  \Description{Performance comparison on both closed-ended and open-ended QA across multiple LVLMs.}
  \label{fig:figure5_appendix}
\end{figure}

\subsection{Semantic Attribute Extraction} 

A core challenge in medical VQA is bridging the gap between low-level pixel cues and high-level linguistic reasoning. 
To address this, we implement an annotation-based semantic conversion pipeline inspired by Vepa et al.~\cite{murari2025multimodal}.

We translate pixel-level tumor masks into structured semantic attributes via a deterministic geometric mapping. 
Specifically, we extract morphology, localization, and spatial spread descriptors as follows.

\paragraph{Morphology.}
We quantify lesion shape compactness using the circularity metric:
\begin{equation}
\small
C = \frac{4\pi A}{P^2},
\end{equation}
where $A$ and $P$ denote the lesion area and perimeter, respectively. 
Lower values of $C$ indicate increased shape irregularity.

\paragraph{Localization.}
The lesion centroid is computed from spatial image moments as:
\begin{equation}
\small
C_x = \frac{M_{1,0}}{M_{0,0}}, \quad 
C_y = \frac{M_{0,1}}{M_{0,0}},
\end{equation}
where the $(p,q)$-th order moment is defined as
\begin{equation}
\small
M_{pq} = \sum_x \sum_y x^p y^q I(x,y),
\end{equation}
with $I(x,y)$ denoting the binary lesion mask.
This descriptor encodes the spatial position of the lesion within the imaging plane.

\paragraph{Spatial Spread.}
To capture lesion multifocality, we define the dominant component ratio:
\begin{equation}
\small
f_{core} = \frac{A_{\max}}{\sum_i A_i},
\end{equation}
where $A_i$ denotes the area of the $i$-th connected component and $A_{\max}$ corresponds to the largest one. 
Lower values indicate increased spatial dispersion.

Based on standardized thresholds (e.g., $C < 0.5$ for irregular morphology), these metrics yield quantitative descriptors of tumor shape, location, and spread. 
Importantly, these structured attributes serve as verifiable intermediate evidence, grounding the model’s reasoning in physical image properties rather than statistical hallucinations.

Building on these attributes, we synthesize natural language descriptions by integrating imaging modalities, tumor subtypes, and the extracted geometric features. 
Crucially, we adopt an explicit indication strategy for missing metadata: instead of omitting unavailable fields, we explicitly mark them as ``unknown.'' 
This design guides the downstream multimodal model to acknowledge data gaps and actively infer missing diagnostic cues from visual evidence. 
Ultimately, this pipeline establishes a reliable mapping from $\text{Pixel Masks} \rightarrow \text{Semantic Evidence} \rightarrow \text{Textual Prompts}$, forming the foundational input for subsequent LLM-based reasoning.The complete mapping rules and implementation details are provided in the Appendix.

\begin{figure*}[t!] 
  \centering
  \includegraphics[width=0.95\textwidth]{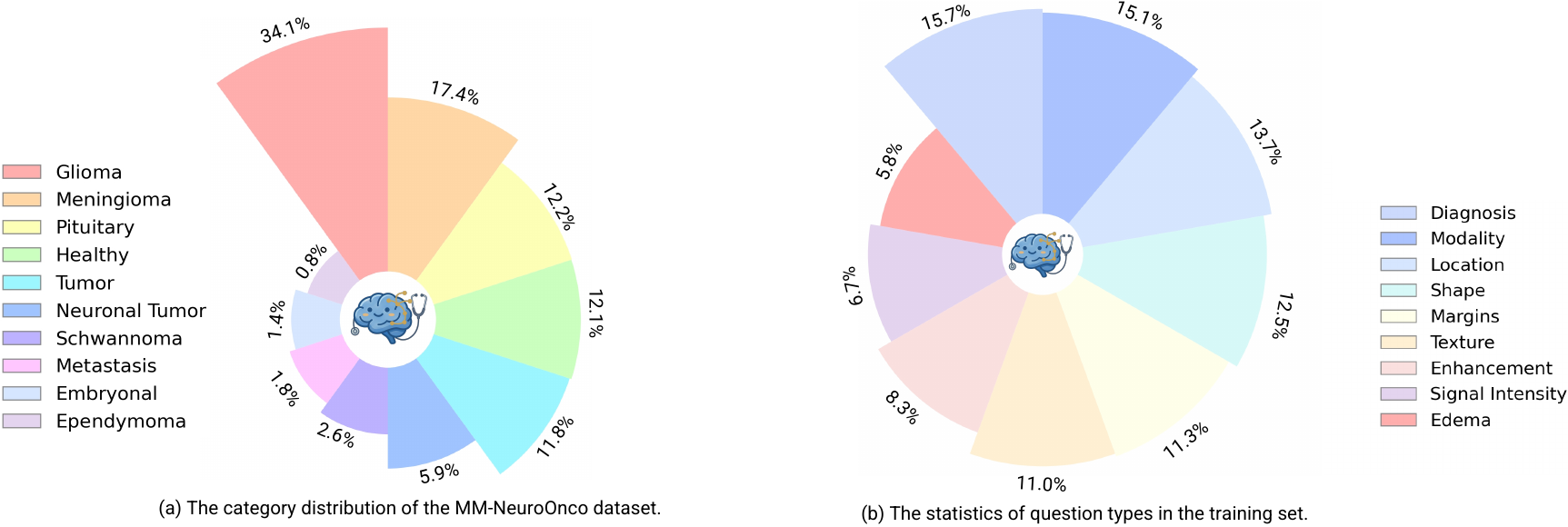}
  \caption{Statistical distributions of the MM-NeuroOnco dataset.}
  \Description{Enjoying the baseball game from the third-base seats. Ichiro Suzuki preparing to bat.}
  \label{fig:figure3}
\end{figure*}

\begin{figure}[t]
  \centering
  \includegraphics[width=0.9\columnwidth]{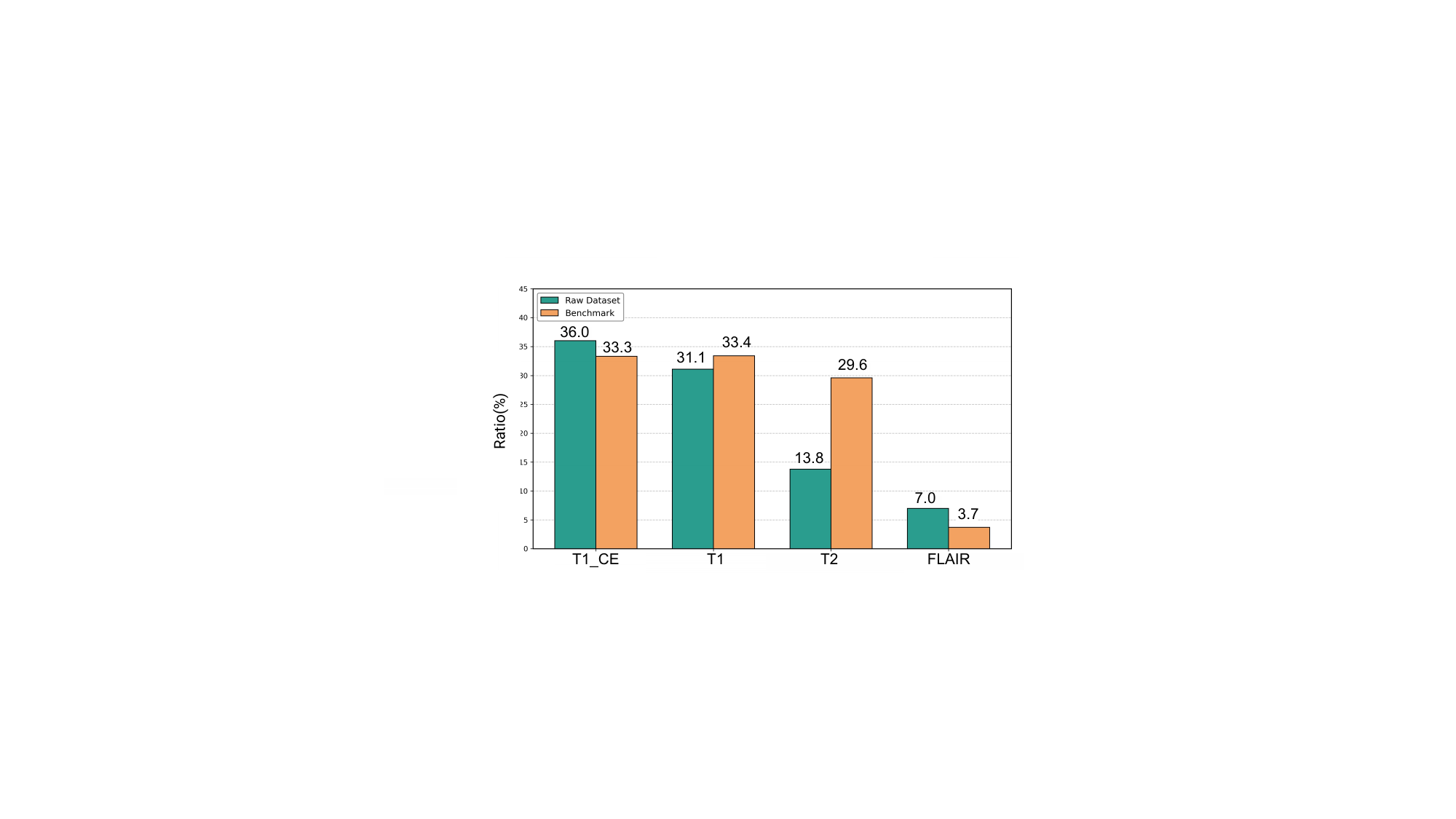}
  \caption{Modality distribution of the MM-NeuroOnco dataset and benchmark.The figure compares the proportional distributions of different MRI modalities in the dataset and the benchmark.}
  \Description{An illustration for Figure 4.}
  \label{fig:figure4}
\end{figure}

\subsection{Multi-Model Refinement \& Quality Control}
While segmentation masks are available for some samples, raw multi-source MRI datasets severely lack the descriptive radiological attributes essential for clinical reasoning---such as enhancement patterns, edema severity, and intratumoral textures. To generate reliable ``silver-standard'' annotations at scale, we devise an automated Multi-Model Collaborative Pipeline, as illustrated in Stage~III of Figure~\ref{fig:figure2}. This framework operates on a ``Dual-Path Extraction -- Consensus Fusion -- Visual Verification'' mechanism, explicitly designed to mitigate hallucinations through cross-validation among heterogeneous large models.

\noindent\textbf{Stage 1: Dual-Path Independent Extraction.} We leverage two heterogeneous and powerful commercial Vision-Language Models (VLMs) to independently analyze each MRI slice and extract medical semantics into a unified structured format.

\noindent\textbf{Stage 2: Consensus-Based Fusion.} We apply a rigorous field-level fusion strategy to merge the dual outputs. Fields with exact matches are directly retained. For minor discrepancies in degree descriptions, we adopt a downgrade strategy: for instance, conflicts such as ``mild'' vs. ``moderate'' are unified into a broader label of ``Present.'' Conversely, significant semantic conflicts are set to null, thereby maximizing the reduction of hallucination risks introduced by large models.

\noindent\textbf{Stage 3: Visual Final Review and Quality Control.} To further intercept subtle errors, we introduce a third high-performance visual model to conduct a final review of the candidate labels. This stage strictly enforces a ``subtraction-only'' mechanism: the model is permitted only to reject untrustworthy fields or discard high-risk samples entirely, but is strictly prohibited from adding new information. This design blocks the propagation of hallucinations, ensuring the high confidence of the final data.

Through this pipeline, we bridge the semantic gap in the raw data and extract structured medical evidence, forming the foundation of our multimodal instruction dataset. Detailed extraction prompts and explanations are provided in the Appendix.

\subsection{Instruction Data Construction}

Our instruction generation pipeline is fundamentally grounded in verifiable supervision signals rather than open-ended generation. We anchor the process in two reliable data sources: (i) Golden Labels (tumor type, modality, segmentation masks) provided by the original datasets; and (ii) Silver Labels (structured radiological attributes) derived from our multi-model pipeline. By treating these labels as immutable facts, we ensure that the subsequent CoT reasoning is evidence-based, effectively mitigating the hallucinations common in medical text generation. 
Furthermore, the category distribution of the dataset and the distribution of question types are illustrated in Figure~\ref{fig:figure3}(a) and Figure~\ref{fig:figure3}(b), respectively.

We employ a locally deployed Qwen3-Next-80B to synthesize these labels into VQA instances.Adversarial Construction for Closed-Ended QA: For multiple-choice tasks, we leverage the LLM's medical knowledge to construct diverse questions and hard negatives---distractors that are visually or pathologically similar to the ground truth (e.g., confusing glioblastoma with solitary metastasis due to similar ring enhancement). This adversarial design prevents models from achieving artificially high scores through simple elimination shortcuts.

Evidence-Driven CoT: To construct high-quality CoT, we synthesize the structured silver attributes with deterministic geometric features (e.g., location, size) calculated from segmentation masks. By explicitly injecting these verified facts into the reasoning path, we force the model to mimic the standardized clinical workflow: identifying modality $\rightarrow$ localizing the lesion $\rightarrow$ analyzing morphology $\rightarrow$ deducing the pathology. This approach ensures that the generated reasoning is grounded in objective evidence rather than open-ended hallucination.

\section{MM-NeuroOnco Dataset Analysis}

\subsection{Instruction Dataset Statistics}

We applied our multi-model semantic pipeline to process over 20,000 samples. 
During sampling, instead of enforcing artificial class balance, we deliberately preserved the authentic epidemiological distribution of brain tumors. 
As shown in Figure~\ref{fig:figure3} (a), the dataset generally aligns with real-world clinical incidence rates~\cite{bray2024global}. 
By retaining this long-tail distribution, we enable models to learn robust representations that reflect the true prevalence priors encountered in complex clinical scenarios.

Following post-processing—including downgrade mapping and structural flattening—we obtained a semantically rich dataset featuring an average of 2.4 fine-grained medical attributes (e.g., enhancement patterns, edema) per sample.

\subsection{Benchmark Composition}

To establish a rigorous evaluation standard, we constructed MM-NeuroOnco-Bench by independently sampling 1,000 images from the raw pool; its modality distribution is illustrated in Figure~\ref{fig:figure4}.

\textbf{Ground Truth Integrity.} Unlike the training set, we enforced a strict ``No-LLM-Inference'' policy for the benchmark's ground truth. All attribute labels are derived exclusively via human annotation or mask mapping. We explicitly excluded any predictions or completions from large models, thereby fundamentally eliminating evaluation bias stemming from model hallucinations.

\textbf{Question Diversity.} While the ground truth is immutable, we utilized a local Qwen3-Next-80B to diversify the linguistic phrasing of questions. The resulting benchmark comprises over 2,000 closed-ended and 1,000 open-ended questions.These questions comprehensively cover multidimensional clinical features, providing a robust yardstick for assessing specialized medical MLLMs.

\begin{table*}[t!]
\centering
\caption{Overall performance comparison on the closed-ended benchmark.
Best results are highlighted in \textbf{bold}, and second-best results are \underline{underlined}.
Colored deltas indicate performance changes introduced by Chain-of-Thought (CoT) supervision.
Qwen3-VL-8B is evaluated in a zero-shot setting as an open-source baseline.}
\label{tab:main_results}

\footnotesize
\renewcommand{\arraystretch}{1.12}
\setlength{\tabcolsep}{5.5pt}

\begin{tabularx}{\textwidth}{p{5.0cm}|YYYYYY}
\toprule
\textbf{Model} & Diagnosis & Size & Shape & Spread & Location & \textbf{Overall} \\
\midrule

\grouprow{General-Purpose LVLMs}
\addlinespace[0.15em]
GPT-5.1
& 37.0 & 32.3 & 21.2 & \underline{58.0} & 38.1 & 37.2 \\
Gemini-3-Flash
& \underline{41.9} & 29.8 & 30.8 & 57.9 & 41.8 & \underline{40.9} \\
Claude-Sonnet-4.0
& 33.2 & 34.5 & 28.3 & 54.4 & 35.5 & 35.9 \\
Qwen3-VL-8B
& 24.3 & 31.6 & 30.0 & 47.2 & 30.6 & 30.1 \\
\midrule

\grouprow{Medical Specialist LVLMs}
\addlinespace[0.15em]
HuLuMed-32B
& 38.2 & 27.7 & 32.9 & 52.1 & 33.2 & 37.3 \\
Lingshu-7B
& 39.2 & 30.0 & 32.9 & 45.0 & 37.5 & 37.6 \\
Lingshu-32B
& 36.0 & 28.3 & 26.1 & 52.8 & 33.9 & 35.6 \\
HuLuMed-7B
& 36.4 & 30.3 & \textbf{40.7} & 28.0 & 29.0 & 34.0 \\
MedGemma-27B
& 33.1 & 29.0 & 32.9 & 33.2 & 27.7 & 31.8 \\
MedGemma-1.5-4B
& 26.9 & 28.0 & 28.0 & 23.5 & 25.1 & 26.5 \\
\midrule

\grouprow{Ours (Domain-Adapted LVLMs)}
\addlinespace[0.15em]
\textsc{NeuroOnco-GPT} {\scriptsize\textnormal{\textcolor{blue}{[Ours]}}}
& 40.6 & \underline{42.0} & 27.7 & 34.5 & \textbf{53.8} & 40.0 \\
\textsc{NeuroOnco-GPT} (CoT) {\scriptsize\textnormal{\textcolor{blue}{[Ours]}}}
& \textbf{51.4} \pos{10.8}
& \textbf{42.4} \pos{0.4}
& \underline{40.4} \pos{12.7}
& \textbf{70.4} \pos{35.9}
& \underline{52.1} \dneg{1.7}
& \textbf{51.4} \posbold{11.4} \\
\bottomrule
\end{tabularx}
\end{table*}

\begin{table}[t]
\centering
\caption{Ablation study on the rejection mechanism.
N denotes the standard option setting, while R denotes the setting with an explicit rejection mechanism.
Colored deltas indicate performance changes relative to the 4-N setting.}
\label{tab:hard-negative-ablation}

\small
\renewcommand{\arraystretch}{1.25}
\setlength{\tabcolsep}{10pt}

\begin{tabular}{lccc}
\toprule
\textbf{Model} & \textbf{4-N} & \textbf{5-N} & \textbf{5-R} \\
\midrule
HuLuMed-32B
& 59.56
& 56.90 \dneg{2.66}
& 49.40 \dneg{10.16} \\

Lingshu-32B
& 57.21
& 55.77 \dneg{1.44}
& 46.93 \dneg{10.28} \\

GPT-5.1
& 59.03
& 56.74 \dneg{2.29}
& 49.62 \dneg{9.41} \\

\midrule
\textbf{Average}
& \textbf{58.60}
& \textbf{56.47} \dneg{2.13}
& \textbf{48.65} \dneg{9.95} \\
\bottomrule
\end{tabular}
\end{table}

\section{MM-NeuroOnco Bench}

\subsection{Rejection-Aware Evaluation Strategy}

Existing medical VQA benchmarks often suffer from naive distractor construction. Distractors frequently exhibit obvious visual or anatomical discrepancies from the ground truth (e.g., pairing a Brain Tumor query with Lung Nodule options). This allows models to rely on "Shortcut Learning"---solving questions via simple elimination rather than genuine pathological comprehension. Moreover, traditional evaluations adhere to a "Forced-Choice Paradigm," presupposing that the correct answer is invariably present. This fundamentally contradicts real-world clinical practice, where diagnosis is a rigorous process of hypothesis exclusion, often requiring a physician to withhold judgment when evidence is insufficient.

To simulate this authentic decision-making environment, we introduce a Rejection-Aware Evaluation Strategy. We uniformly append a fifth option---"None of the above"---to every closed-ended question. This simple mechanism forces the model to verify its internal knowledge against the provided options rather than exploiting probability distributions.

As shown in Table~2, the experimental results indicate that this mechanism leads to a $\sim$10\% drop in accuracy for Large Language Models compared to the standard four-option setting.
To confirm this decline stems from increased difficulty rather than just option quantity, we conducted rigorous ablation studies. Even compared to a five-option setting with conventional distractors, our refusal strategy induces an additional 8\% performance drop. This gap indicates that previous "State-of-the-Art" scores were inflated by test-taking tricks. By forcing models to scrutinize their own knowledge boundaries, this protocol offers a far more objective yardstick for diagnostic evaluation.

\begin{figure*}[t]
  \centering
  \includegraphics[width=0.99\textwidth]{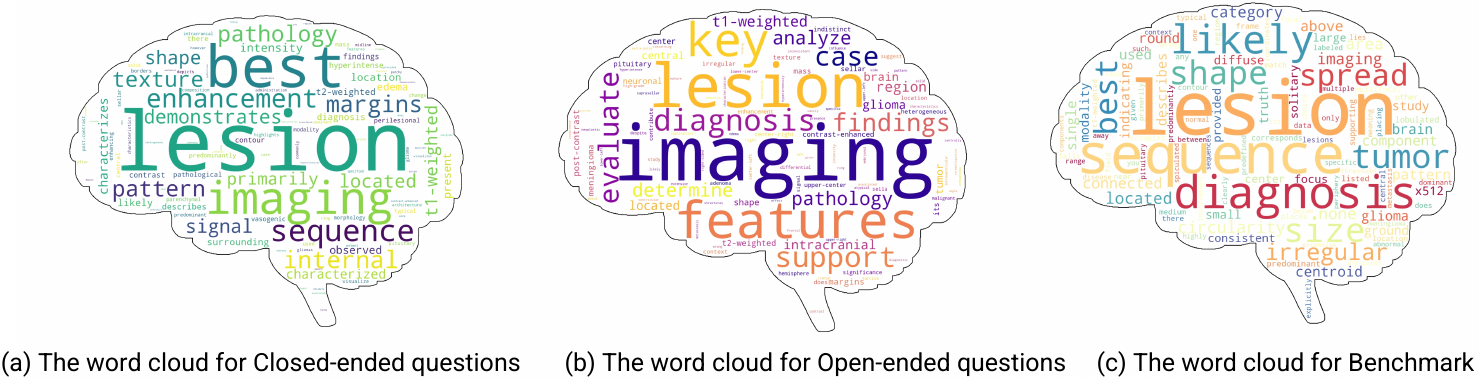}
  \caption{Word cloud visualizations illustrating the linguistic distributions of (a) closed-ended questions, (b) open-ended questions, and (c) the evaluation benchmark.}
  \Description{Overall pipeline and system overview of the proposed framework.}
  \label{fig:figure6}
\end{figure*}

\subsection{Evaluation Metrics}

We adopt two evaluation metrics tailored for closed-ended and open-ended questions, respectively.For Closed-Ended Questions, we use Accuracy as the standard evaluation metric.

For Open-Ended Questions, following previous works~\cite{hao2025oralgpt,hao2025towards}, we construct a specialized prompt and leverage a locally deployed Qwen3-80B-Instruct to assist with the evaluation. Within the prompt, we incorporate a Hierarchical Scoring Rubric that assigns a score ranging from 0 to 10 based on each sample's input question, ground truth, and model output. Crucially, to ensure clinical rigor, this rubric enforces a penalty mechanism that strictly caps scores for critical safety errors (e.g., laterality confusion or hallucinations). The full details of the evaluation prompt can be found in the supplementary materials.

\section{Experiments}
\label{sec:experiments}

\subsection{Experimental Setup}

To delineate the capability boundaries of current technology, we benchmark 10 representative LVLMs, stratified into two distinct clusters: (1) General-Purpose Giants: we select state-of-the-art closed-source models, GPT-5.1, Gemini-3-Flash, and Claude-Sonnet-4.0, representing the current pinnacle of general multimodal reasoning; and (2) Medical Specialists: we include open-source vertical domain models ranging from 1.5B to 32B parameters, such as HuLuMed, Lingshu, and MedGemma. These models were continually pre-trained or fine-tuned for biomedical tasks.

Implementation Protocols. Closed-source models are accessed via official APIs. Open-source models are evaluated on a server node with $4 \times$ NVIDIA H100 (80G) GPUs. For evaluation, we use Accuracy as the evaluation metric for closed-ended tasks. For open-ended inquiries, we employ the LLM-as-a-Judge paradigm described in Section~4.2, using Qwen3-80B-Instruct as the impartial judge.

Validation via SFT. To demonstrate the efficacy of our instruction data, we perform Supervised Fine-Tuning (SFT) on a Qwen3-VL-8B base model. We utilize the LLaMA-Factory framework with a LoRA strategy. All hyperparameters are kept at their default values, and the training is conducted for a single epoch. The resulting model is named \textbf{NeuroOnco-GPT}.

\subsection{Evaluation Results}

Based on the overall performance on MM-NeuroOnco-Bench (Table~\ref{tab:main_results}), together with the rejection ablation study (Table~\ref{tab:hard-negative-ablation}), we summarize three key insights into the current state of neuro-oncology AI:

\begin{figure}[t]
  \centering
  \includegraphics[width=0.99\columnwidth]{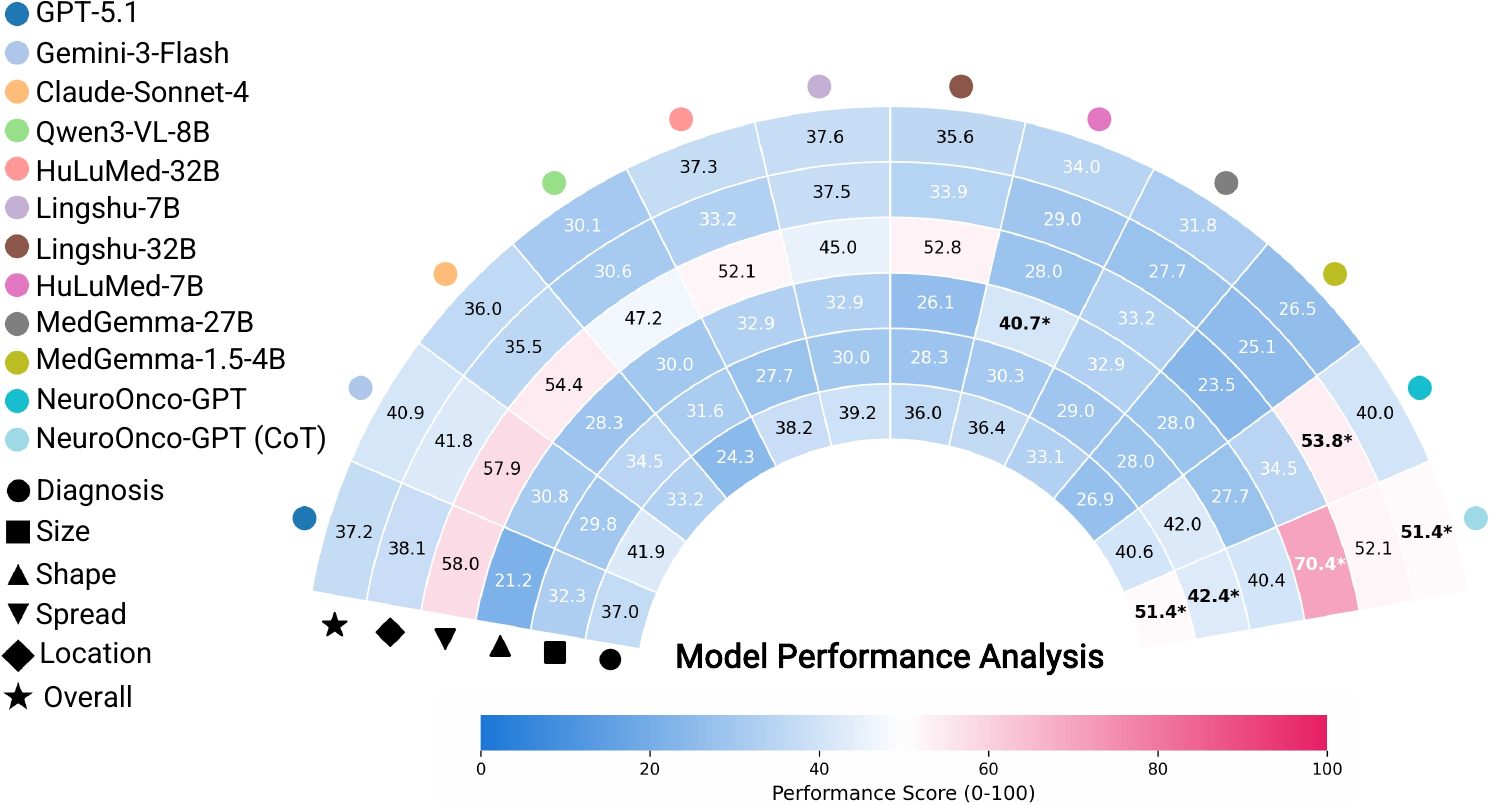}
  \caption{Heatmap comparing the performance of multiple LVLMs on closed-ended QA tasks across different dimensions.}
  \Description{Performance comparison on both closed-ended and open-ended QA across multiple LVLMs.}
  \label{fig:figure5_appendix}
\end{figure}

\begin{itemize}
    \item \textbf{General Intelligence is Not Enough.}  
    MM-NeuroOnco-Bench proves to be a challenging benchmark for existing LVLMs. Even Gemini-3-Flash, the strongest commercial general-purpose model in our evaluation, achieves only \textbf{40.9\%} overall accuracy. This result indicates that brain tumor diagnosis---which requires integrating subtle visual cues with structured anatomical reasoning---remains a difficult problem that is not adequately addressed by general-purpose multimodal intelligence alone.

    \item \textbf{The ``Medical'' Label Does Not Guarantee Superior Performance.}  
    Surprisingly, medical-specialized LVLMs do not exhibit a consistent advantage over general-purpose models. For example, the best-performing specialist, HuLuMed-32B (37.3\%), as well as Lingshu-7B (37.6\%), remain notably behind Gemini-3-Flash (40.9\%). This observation suggests that simply exposing models to biomedical data is insufficient when such data lacks the semantic density and explicit reasoning structure required for reliable MRI interpretation.

    \item \textbf{Rejection-Aware Evaluation Increases Task Difficulty and Reveals Capability Boundaries.}  
    Traditional multiple-choice evaluation settings may partially obscure model uncertainty by implicitly assuming that a correct answer always exists. Under such settings, models can often achieve relatively high scores through the elimination of implausible options, without fully grounding their decisions in imaging evidence. After introducing an explicit rejection option, the average accuracy drops by an additional \textbf{7.82\%}, indicating a clear increase in task difficulty. This result suggests that rejection-aware evaluation provides a more conservative and informative perspective for assessing model capability boundaries in scenarios that more closely resemble real-world clinical decision-making.
\end{itemize}

\begin{figure*}[t]
  \centering
  \includegraphics[width=0.9\textwidth]{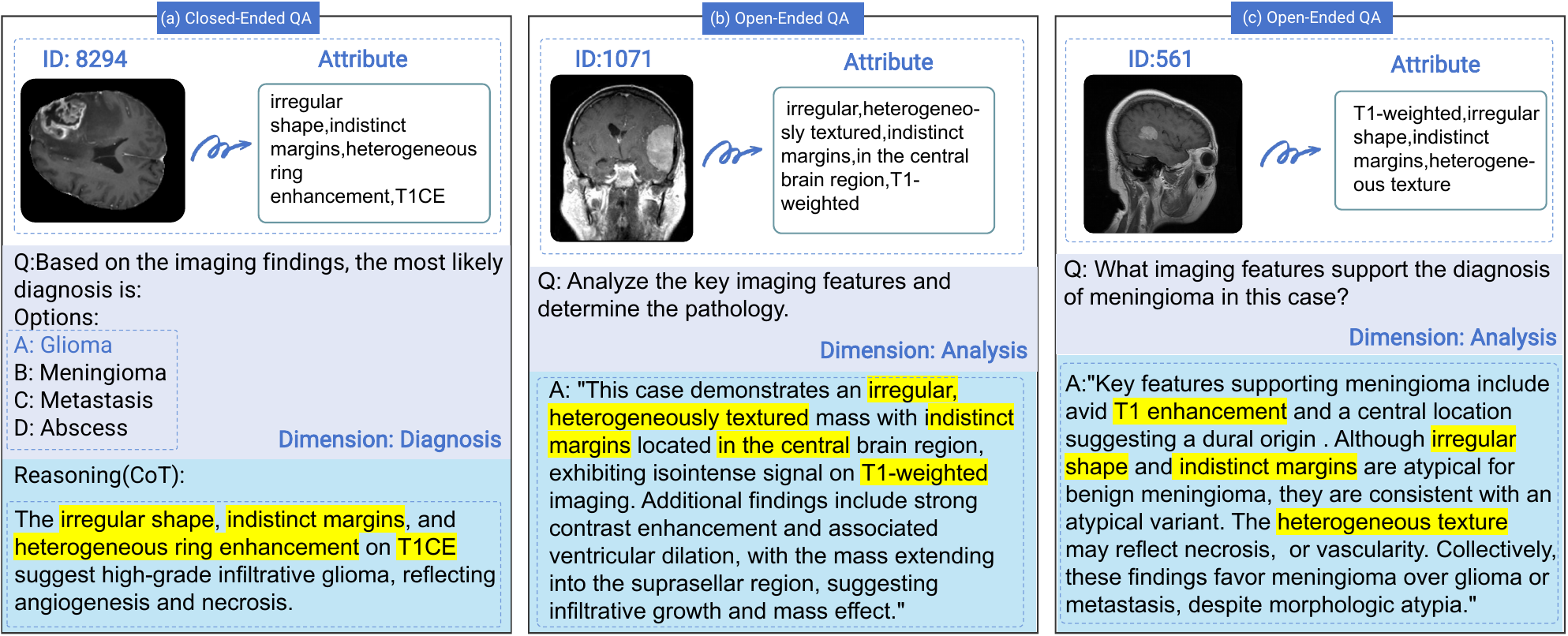}
  \caption{Three representative case studies for closed-ended and open-ended QA.}
  \Description{Supplementary visualization for the data construction pipeline.}
  \label{fig:figure7_appendix}
\end{figure*}

\subsection{Ablation Study}

We conduct two targeted ablation studies to examine the rigor of our evaluation protocol and the impact of reasoning-oriented supervision.

\textbf{Distinguishing Comprehension from Guessing.}  
To characterize model behavior under different evaluation settings, we compare three configurations (Table~\ref{tab:hard-negative-ablation}): a standard four-choice setting, a five-choice setting with a conventional distractor, and a five-choice setting with an explicit rejection option. Results show that adding a conventional distractor leads to only a minor performance decrease ($-2.13\%$), whereas introducing the rejection mechanism results in a larger accuracy drop of $9.95\%$, corresponding to an additional $7.82\%$ decrease relative to the normal five-choice setting. These findings indicate that when models are required to explicitly consider the possibility that none of the provided options is correct, the evaluation becomes more challenging and helps to better characterize model capability boundaries in diagnostic tasks.

\textbf{The Effect of CoT Supervision.} 
We further evaluate the role of CoT data in a low-resource fine-tuning setting (Table~\ref{tab:hard-negative-ablation}). Compared to fine-tuning with instruction data that does not include CoT, incorporating explicit CoT reasoning paths leads to a further performance improvement, with an overall gain of $11.4\%$. This result indicates that guiding models through a structured clinical workflow (Observation $\rightarrow$ Reasoning $\rightarrow$ Decision) is more effective than training them to memorize shallow image--label associations.

\subsection{Case Study}

Figure~\ref{fig:figure7_appendix} presents three representative case studies from our instruction dataset. 
In the closed-ended example (Figure~\ref{fig:figure7_appendix}a), each sample includes extracted medical attributes, LLM-constructed questions and options, as well as an attribute-based CoT that explicitly articulates the diagnostic logic. 
For open-ended tasks (Figure~\ref{fig:figure7_appendix}b and Figure~\ref{fig:figure7_appendix}c), the pipeline generates QA pairs in which the answers are logically synthesized from the underlying medical attributes. 
These examples demonstrate that our pipeline effectively transforms discrete metadata into structured, hallucination-free instructions, providing high-quality supervision signals for model training.

\section{Conclusion}

In this work, we introduce \textbf{MM-NeuroOnco}, a large-scale multimodal instruction dataset and evaluation benchmark designed to enrich fine-grained medical semantics and mitigate the long-tail distribution issues inherent in brain tumor MRI data. To overcome the prohibitive costs of manual annotation, we devise an automated Multi-Model Collaborative Pipeline that supplements raw scans with verifiable diagnostic attributes, effectively transforming pixel-level data into evidence-driven reasoning chains. Building on this foundation, we further develop \textbf{NeuroOnco-GPT} and validate its effectiveness through a rigorous Rejection-Aware Evaluation strategy. Our experimental results show that this mechanism can, to a certain extent, alleviate the ``illusion of competence'' induced by shortcut learning, thereby providing a more objective reference for assessing the capability boundaries of multimodal large language models in specialized medical domains.

\section{Limitations and Ethical Considerations}

While this work provides a systematic benchmark and dataset for neuro-oncology VQA, several limitations remain.First, the current benchmark is primarily constructed based on single 2D MRI slices. Although this setting is effective for validating fundamental pattern recognition capabilities, it lacks the volumetric context that is essential in real-world clinical practice. Given the substantial computational overhead associated with full 3D models, future work will explore a \emph{multi-slice joint diagnosis} paradigm, enabling models to collaboratively reason over a compact set of key slices. This design aims to better approximate the radiologist’s workflow while avoiding the redundancy and cost of full volumetric processing.Second, some diagnostic semantics in MM-NeuroOnco rely on automated data generation pipelines. Despite careful quality control, the resulting attributes may still be affected by noise originating from imperfect image quality or inaccuracies in the original annotations of public datasets. Addressing these issues will require more robust data curation strategies and continued refinement of the semantic extraction process.

From an ethical perspective, all data used in this study are sourced from publicly available brain tumor MRI datasets, following data usage paradigms consistent with established benchmarks such as BraTS. These datasets are released in a de-identified form by their original providers and contain no personally identifiable or protected health information. All samples used in this work are fully traceable to their original sources, and their usage strictly complies with the corresponding data usage agreements and licenses.


\newpage

\bibliographystyle{ACM-Reference-Format}
\nocite{*}
\bibliography{references}

\onecolumn 
\appendix

\begin{center}
    {\huge \textbf{MM-NeuroOnco: A Multimodal Benchmark and Instruction \\ Dataset for MRI-based Brain Tumor Diagnosis}} \\[20pt] 
    {\huge \textbf{Supplementary Material}}
\end{center}

\vspace{1.0cm} 

\noindent {\Large \textbf{Contents}}

\vspace{0.5cm} 
\setlength{\parskip}{0pt} 

\noindent
\textbf{A\quad Details on Training Data Construction} \hfill \textbf{12} \\
\vspace{3pt} 
\noindent \hspace*{2em} A.1\quad Multi-Source Data Aggregation and Standardization \dotfill 12 \\
\vspace{3pt}
\noindent \hspace*{2em} A.2\quad Medical Semantic Mapping \dotfill 12 \\
\vspace{3pt}
\noindent \hspace*{2em} A.3\quad Automated Coarse Description Generation \dotfill 12 \\
\vspace{3pt}
\noindent \hspace*{2em} A.4\quad Dual-MLLM Visual Sign Extraction Framework \dotfill 13 \\
\vspace{3pt}
\noindent \hspace*{2em} A.5\quad Train–Benchmark Split Protocol \dotfill 14 \\
\vspace{10pt}

\noindent
\textbf{B\quad Process Monitoring and Hallucination Mitigation} \hfill \textbf{14} \\
\vspace{3pt}
\noindent \hspace*{2em} B.1\quad Process Monitoring via Average Information Rate (AIR) \dotfill 14 \\
\vspace{3pt}
\noindent \hspace*{2em} B.2\quad Hallucination Mitigation via Multi-Model Cross-Validation \dotfill 15 \\
\vspace{3pt}
\noindent \hspace*{2em} B.3\quad Silver Annotation Quality Audit \dotfill 15 \\
\vspace{10pt}

\noindent
\textbf{C\quad Task and Evaluation Design} \hfill \textbf{16} \\
\vspace{3pt}
\noindent \hspace*{2em} C.1\quad Automated QA Generation \dotfill 16 \\
\vspace{3pt}
\noindent \hspace*{2em} C.2\quad Rationale for Feature Selection and Clinical Relevance \dotfill 16 \\
\vspace{3pt}
\noindent \hspace*{2em} C.3\quad Open-Ended Evaluation: LLM-as-a-Judge Protocol \dotfill 17 \\
\vspace{10pt}

\noindent
\textbf{D\quad Supplementary Implementation and
Qualitative Analysis} \hfill \textbf{17} \\
\vspace{3pt}
\noindent \hspace*{2em} D.1\quad Supplementary Figures \dotfill 17 \\
\vspace{3pt}
\vspace{10pt}

\vspace{1.5cm} 

\twocolumn


\section*{A\quad Details on Training Data Construction}
\label{sec:data_construction}

\noindent
\textbf{A.1\quad Multi-Source Data Aggregation and Standardization.}

To establish a comprehensive and clinically representative benchmark, we aggregated 20 distinct datasets, spanning diverse samples from multi-center challenges (e.g., BraTS) to open-source repositories. The specific sources and distribution of the datasets are detailed in Table~\ref{tab:data_sources} and Figure~\ref{fig:supp_fig9}.Confronting the significant heterogeneity in storage formats across these sources---encompassing 3D NIfTI volumes, 2D DICOM sequences, and various non-standard annotation formats such as YOLO/XML---we engineered seven specialized data processing scripts. These pipelines systematically executed directory parsing, modality alignment, and annotation conversion, with a particular focus on rasterizing vector-based YOLO normalized coordinates into unified binary masks.

After standardization, we conducted a large-scale duplicate removal procedure to enhance data integrity. Specifically, pixel-level comparison was performed after unified preprocessing, and MRI slices with identical pixel matrices were considered duplicates. This process eliminated approximately 29,000 redundant images, reducing the dataset size from over 100,000 raw slices to more than 70,000 unique MRI slices. All duplicate removal was conducted globally across the aggregated data pool prior to dataset partitioning to prevent cross-source redundancy.Through this standardization and integrity filtering workflow, we transformed a disordered, multi-source, and heterogeneous brain tumor dataset into a unified, structured index optimized for downstream multimodal instruction construction and benchmark evaluation.

\vspace{8pt}

\begin{figure}[!ht]
  \centering
  \includegraphics[width=0.9\columnwidth]{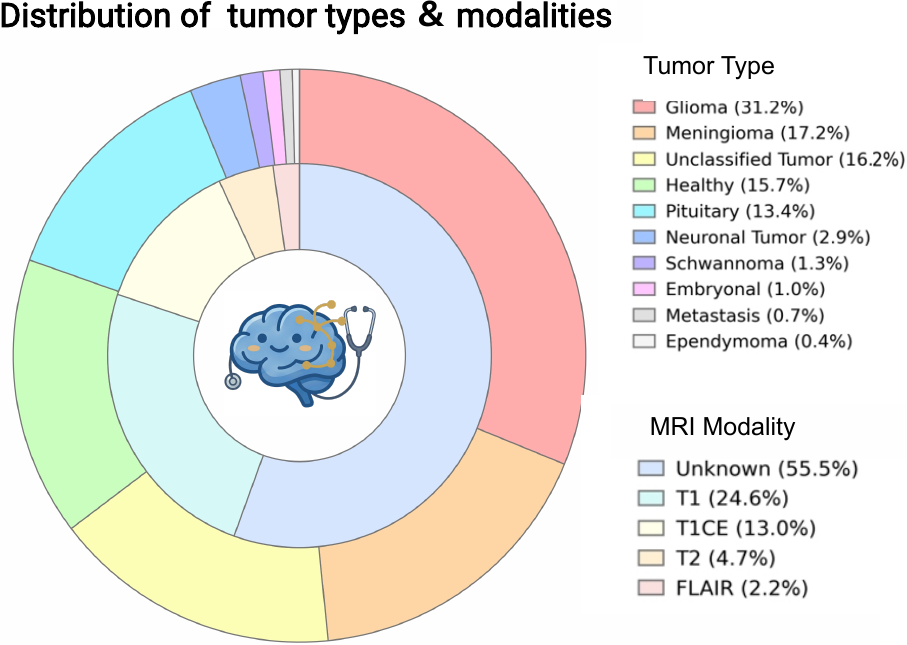}
  \caption{Distribution of tumor types and modalities.The outer ring shows the distribution of brain tumor types, while the inner ring displays the distribution of MRI modalities.}
  \Description{Supplementary Figure 9.}
  \label{fig:supp_fig9}
\end{figure}

\noindent
\textbf{A.2\quad Medical Semantic Mapping.}

Since the aggregated datasets comprise diverse, unregistered MRI scan planes (e.g., axial, coronal, sagittal), traditional anatomical localization is prone to ambiguity. To address this, we developed a \textit{Purely Visual Morphological Pipeline} that maps binary masks into structured semantic labels across four distinct dimensions:

\begin{enumerate}[leftmargin=*] 
    
    \item \textbf{Relative Tumor Size.} 
    To quantify tumor burden, we calculate the relative area ratio $R_{\text{area}}$:
    \begin{equation}
        R_{\text{area}} = \frac{N_{\text{tumor}}}{H \times W}
    \end{equation}
    Based on empirical clinical thresholds, lesions are categorized as:
    \begin{itemize}
        \item \textbf{Small/Focal:} $R_{\text{area}} < 1\%$
        \item \textbf{Medium:} $1\% \le R_{\text{area}} < 5\%$
        \item \textbf{Large/Extensive:} $R_{\text{area}} \ge 5\%$
    \end{itemize}

    \item \textbf{Morphological Characterization.} 
    Adhering to the IBSI guidelines, we employ Circularity ($C$) and Elongation ($E$) to characterize invasive features.
    
    $C$ measures the deviation from a perfect circle:
    \begin{equation}
        C = \frac{4\pi A}{P^2}
    \end{equation}
    
    $E$ represents the ratio of major to minor axes derived via PCA:
    \begin{equation}
        E = \frac{\lambda_{\text{major}}}{\lambda_{\text{minor}}}
    \end{equation}
    
    \noindent\textit{Classification Logic:}
    \begin{itemize}
        \item \textbf{Irregular/Spiculated:} $C < 0.5$
        \item \textbf{Round/Oval:} $C \ge 0.8$ and $E \le 1.5$
        \item \textbf{Lobulated:} Intermediate cases
    \end{itemize}

    \item \textbf{Spread and Multiplicity.} 
    We utilize Connected Component Analysis (CCA) to identify the number of independent lesion components ($N_c$) and the core dominance ratio $f_{\text{core}}$:
    \begin{equation}
        f_{\text{core}} = \frac{A_{\text{max}}}{\sum A_i}
    \end{equation}
    Classifications include:
    \begin{itemize}
        \item \textbf{Solitary:} $N_c = 1$
        \item \textbf{Dominant with Satellites:} $N_c > 1$ and $f_{\text{core}} \ge 0.7$
        \item \textbf{Scattered/Multifocal:} $N_c > 1$ and $f_{\text{core}} < 0.7$
    \end{itemize}

    \item \textbf{Grid-Based Visual Localization.} 
    We calculate the geometric centroid $(C_x, C_y)$ of the lesion and discretize it into a $3 \times 3$ spatial grid (generating descriptors such as ``Upper-Left'' or ``Center''). This explicit visual coordinate system, independent of anatomical priors, effectively guides the VLM's attention to the pixel distribution of the pathology.

\end{enumerate}

\begin{table*}[t]
\centering
\caption{Inventory of the 20 original data sources incorporated in MM-NeuroOnco.}
\label{tab:data_sources}
\small
\setlength{\tabcolsep}{4pt}
\renewcommand{\arraystretch}{1.12}
\begin{tabularx}{\textwidth}{@{}p{0.75cm} p{4.8cm} R Z p{0.9cm}@{}}
\toprule
\textbf{ID} & \textbf{Dataset Name} & \textbf{Samples} & \textbf{Key Characteristics} & \textbf{Link} \\
\midrule
01 & Brain Tumor Classification (MRI) & 3,264 & 4-class MRI tumor classification (glioma, meningioma, pituitary, no tumor). & \sourcelink{https://www.kaggle.com/datasets/sartajbhuvaji/brain-tumor-classification-mri} \\
02 & Br35H: Brain Tumor Detection 2020 & 1,500 & Binary tumor classification; balanced dataset. & \sourcelink{https://www.kaggle.com/datasets/ahmedhamada0/brain-tumor-detection} \\
03 & Brain Tumor MRI Dataset & 7,023 & 4-class tumor classification; MRI modality unspecified. & \sourcelink{https://www.kaggle.com/datasets/masoudnickparvar/brain-tumor-mri-dataset} \\
04 & Brain Cancer MRI Dataset & 6,056 & Glioma, meningioma, pituitary tumor cases. & \sourcelink{https://www.kaggle.com/datasets/orvile/brain-cancer-mri-dataset} \\
05 & BRISC: Brain Tumor Seg. Dataset & 6,000 & T1-weighted MRI; 4-class labels; segmentation masks. & \sourcelink{https://zenodo.org/records/17524350} \\
06 & Jun Cheng Brain Tumor Dataset & 3,064 & T1-weighted contrast-enhanced MRI; three tumor subtypes. & \sourcelink{https://www.kaggle.com/datasets/mateuszbuda/lgg-mri-segmentation} \\
07 & Figshare Brain Tumor Dataset & 3,000 & T1-weighted contrast-enhanced MRI; benchmark subset. & \sourcelink{https://figshare.com/articles/dataset/brain_tumor_dataset/1512427} \\
08 & Brain Tumor MRI Images (44 Classes) & 4,478 & Multimodal MRI (T1/T1CE/T2); 44 fine-grained tumor classes. & \sourcelink{https://www.kaggle.com/datasets/masoudnickparvar/brain-tumor-mri-images-44-classes} \\
09 & Brain Tumor MRI Images (17 Classes) & 4,449 & Multimodal MRI (T1/T1CE/T2); 17 fine-grained tumor classes. & \sourcelink{https://www.kaggle.com/datasets/masoudnickparvar/brain-tumor-mri-images-17-classes} \\
10 & Brain Tumor Classification 2D & 6,813 & Multimodal MRI (T1/T1CE/T2/FLAIR); pixel-level segmentation masks. & \sourcelink{https://www.kaggle.com/datasets/mehradaria/brain-tumor-classification-2d} \\
11 & Brain Tumor Classification (14 Classes) & 4,455 & 14-class brain tumor subtype classification. & \sourcelink{https://www.kaggle.com/datasets/ahmedhamada0/brain-tumor-classification} \\
12 & Brain Tumor Classification (15 Classes) & 4,288 & 15-class brain tumor subtype classification. & \sourcelink{https://www.kaggle.com/datasets/masoudnickparvar/brain-tumor-classification-15-classes} \\
13 & Brain Tumor Multimodal Image Dataset & 9,618 & Mixed CT and MRI modalities; common tumor types. & \sourcelink{https://www.kaggle.com/datasets/radek1/brain-tumor-multimodal-image} \\
14 & Brain Tumor MRI Dataset for DL & 9,257 & 4-class MRI tumor classification for deep learning. & \sourcelink{https://www.kaggle.com/datasets/masoudnickparvar/brain-tumor-mri-dataset-for-dl} \\
15 & Brain Tumor Segmentation Dataset & 3,064 & T1-weighted contrast-enhanced MRI; binary segmentation masks. & \sourcelink{https://www.kaggle.com/datasets/mateuszbuda/lgg-mri-segmentation} \\
16 & RSNA--MICCAI BraTS 2021 & 8,330 & Multimodal MRI (T1/T1CE/T2/FLAIR); expert-annotated segmentation labels. & \sourcelink{https://www.kaggle.com/competitions/rsna-miccai-brain-tumor-radiogenomic-classification} \\
17 & Brain Tumors 256$\times$256 Dataset & 3,096 & 4-class tumor classification; standardized spatial resolution. & \sourcelink{https://www.kaggle.com/datasets/ahmedhamada0/brain-tumors-256x256} \\
18 & Brain Tumor Dataset (Images and Masks) & 6,129 & Unspecified tumor subtypes; segmentation masks provided. & \sourcelink{https://www.kaggle.com/datasets/mohamedhanyyy/brain-tumor-dataset-mask-images} \\
19 & Medical Image Dataset: Detection & 3,903 & 4-class medical image detection; bounding-box annotations (YOLO). & \sourcelink{https://www.kaggle.com/datasets/andrewmvd/medical-mnist} \\
20 & Brain Tumor MRI Classification 2025 & 5,000 & Binary MRI tumor classification; recent dataset. & \sourcelink{https://www.kaggle.com/datasets/gabrielleyva307/brain-tumor-mri-classification-dataset-2025} \\
\bottomrule
\end{tabularx}
\end{table*}

\noindent
\textbf{A.3\quad Automated Coarse Description Generation.}
\label{app:a3}

To transform the extracted structured metadata into textual formats ingestible by Large Language Models (LLMs), we designed a robust, rule-based Natural Language Generation (NLG) engine. This engine serializes image metadata and semantic attributes into coherent natural language descriptions, strictly adhering to a three-part hierarchical template:
$D = [T_{\text{base}}, T_{\text{pathology}}, T_{\text{morphology}}]$.
The overall generation workflow and template instantiation are illustrated in Figure~\ref{fig:supp_fig10}.First, the generator constructs the Imaging Context ($T_{\text{base}}$) based on modality fields; it specifies the exact sequence when known (e.g., \emph{``a T1-weighted brain MRI scan''}) or falls back to a generic description with an explicit ``unknown'' status when the modality is missing, thereby ensuring factual accuracy. Subsequently, the system appends a Pathology Statement ($T_{\text{pathology}}$): for healthy samples, it outputs \emph{``no visible pathological findings''} and terminates generation; for tumor samples, it dynamically generates diagnostic descriptions ranging from specific (e.g., \emph{``signs of Glioma''}) to generalized (e.g., \emph{``an abnormal mass''}) based on the granularity of the available labels.

Following the baseline diagnosis, the engine assesses the availability of pixel-level annotations to construct Morphological Details ($T_{\text{morphology}}$). For samples containing segmentation masks, the system populates syntactic slots with the size, location, shape, and spread attributes extracted in Appendix A.2, generating fine-grained descriptions such as \emph{``The mass is large and located in the upper-left region.''} Crucially, for the substantial portion of samples lacking mask annotations, the engine automatically generates an explicit disclaimer indicating that morphological details are unavailable. This strategy of knowledge boundary definition prevents the multimodal model from hallucinating features in unlabeled regions during subsequent training, ensuring consistency and credibility of the generated text across varying data quality conditions.

\vspace{8pt}

\begin{figure}[!ht]
  \centering
  \includegraphics[width=\columnwidth]{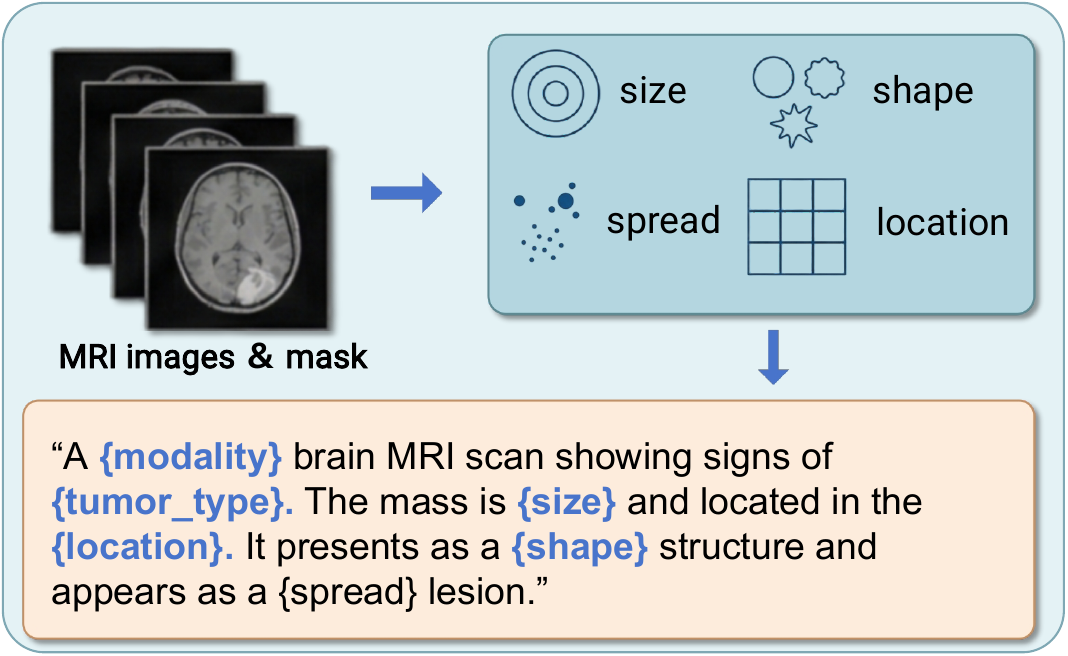}
  \caption{Coarse-grained text description template converted from structured JSON fields into natural language.}
  \Description{Supplementary Figure 10.}
  \label{fig:supp_fig10}
\end{figure}

\begin{figure*}[t]
  \centering
  \includegraphics[width=0.99\textwidth]{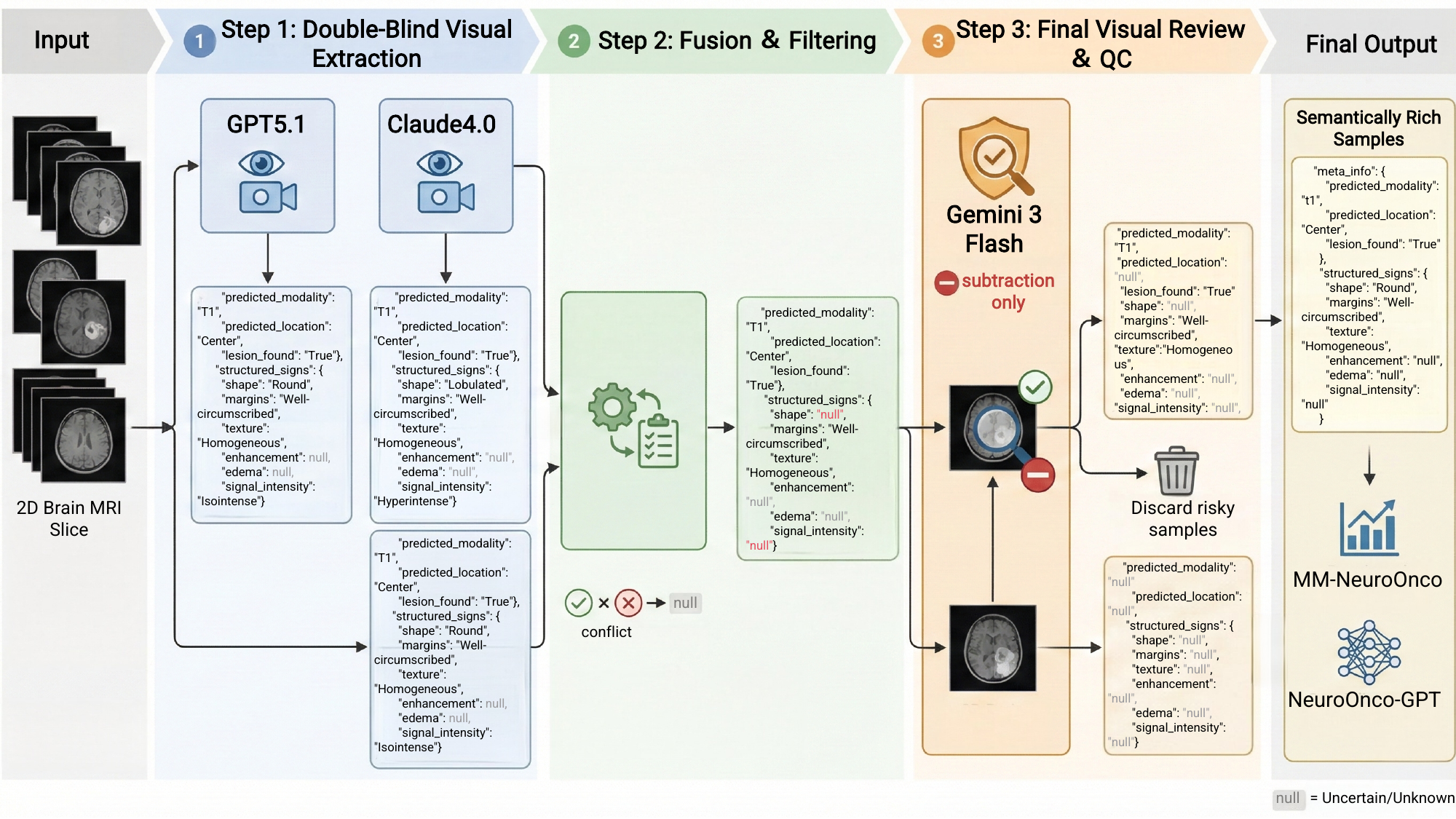}
  \caption{Multi-model Assisted Automated Medical Information Extraction and Quality Control Process.}
  \Description{Supplementary visualization for multi-source data aggregation and standardization.}
  \label{fig:figure8_appendix}
\end{figure*}

\noindent
\textbf{A.4\quad Dual-MLLM Visual Sign Extraction Framework.}
\label{app:a4}

To extract fine-grained radiological signs from 2D MRI slices, we constructed an automated multimodal analysis pipeline that integrates two heterogeneous state-of-the-art models via API services: GPT-5.1 and Claude-Sonnet-4.0. Considering the differences in their underlying architectures and instruction-following behaviors, we applied targeted System Prompt adaptations to improve reasoning consistency and performance. Despite these model-specific adjustments, the overall framework is governed by a unified radiological CoT protocol.

Specifically, both models are instantiated as conservative neuroradiologists adhering to evidence-based principles, with a core ethic of \emph{``Omission over Fabrication''} to mitigate hallucination and ensure that outputs remain grounded in verifiable visual evidence. To maintain reasoning controllability, we enforce a Default Null Policy in which all diagnostic fields are initialized as \texttt{null} and can only be updated after satisfying a structured \textbf{``Location--Appearance--Certainty''}validation (i.e., where the lesion is located, what visual characteristics it presents, and why the conclusion is justified). Additionally, a Pixel Authority Principle prioritizes visual signals over textual priors. To address grayscale ambiguity and modality-specific uncertainty, we further impose domain constraints, including the NAWM baseline (using contralateral normal-appearing white matter as the signal reference) and MRI physics consistency rules (e.g., cerebrospinal fluid must appear hypointense on T1), thereby reducing artifact-induced misinterpretation.

\vspace{8pt}

\noindent\textbf{A.5\quad Train--Benchmark Split Protocol.}
\label{app:a5}

\noindent
To ensure rigorous evaluation and minimize potential data leakage, we establish structured isolation between the instruction dataset and MM-NeuroOnco-Bench. Global pixel-level deduplication is performed prior to partitioning, where MRI slices with identical pixel matrices after unified preprocessing are removed across the aggregated data pool to ensure that no exact duplicate images appear across splits. For volumetric datasets (e.g., brain tumor segmentation benchmarks), only a single representative 2D slice is extracted from each 3D volume, such that each volumetric case contributes at most one sample; this design implicitly enforces volume-level isolation and prevents cross-slice correlations from the same 3D case appearing across training and evaluation sets.

After deduplication and single-slice sampling, we perform stratified partitioning based on tumor categories and imaging modalities to maintain balanced distribution across splits, with detailed statistics shown in Figure~\ref{fig:figure8_appendix}. For datasets originally provided as independent 2D slices without reliable patient- or case-level identifiers, strict patient-level grouping cannot be enforced; in such cases, we rely on pixel-level deduplication combined with stratified partitioning to guarantee slice-level disjointness while preserving distributional balance. Collectively, this protocol ensures that no identical images and no multi-slice correlations from the same volumetric case appear across splits, while maintaining fair and stable benchmarking conditions.

\vspace{8pt}

\begin{table*}[t]
\centering
\caption{Brief description of the four sub-datasets in MM-NeuroOnco and their corresponding data sizes.}
\label{tab:dataset_stats}

\renewcommand{\arraystretch}{1.15}
\renewcommand{\tabularxcolumn}[1]{m{#1}}

\begin{tabularx}{\textwidth}{ c | c | X | c }
\toprule
\textbf{Dataset} & \textbf{Sub-Dataset} & \textbf{Description} & \textbf{Size} \\
\midrule

\multirow{4}{*}{\raisebox{-7\height}{\textbf{MM-NeuroOnco}}}
& Image Pool
& Curated collection of 2D MRI slices from 20 public datasets processed via unified indexing, deduplication, and normalization.
& 73,226 \\ \cmidrule{2-4}

& Attributes
& Integrates 9 LLM-extracted silver labels (e.g., edema and enhancement) alongside human-annotated gold labels such as tumor type.
& 24,726 \\ \cmidrule{2-4}

& VQA Pairs
& Featuring 130k+ closed-ended QA pairs with challenging distractors and CoT reasoning, supplemented by 70k+ open-ended pairs.
& 200k+ \\ \cmidrule{2-4}

& Benchmark
& A high-quality evaluation subset containing 1,000 images annotated with 2k+ closed-ended and 1k+ open-ended questions.
& 3k+ \\
\bottomrule
\end{tabularx}
\end{table*}

\begin{table*}[t!]
\centering
\caption{Overall performance comparison on the open-ended benchmark.
Best results are highlighted in \textbf{bold}, and second-best results are \underline{underlined}.}
\label{tab:open_results}

\footnotesize
\renewcommand{\arraystretch}{1.12}
\setlength{\tabcolsep}{6.0pt}

\begin{tabularx}{\textwidth}{p{5.0cm}|YYYY}
\toprule
\textbf{Model} & Detail & Location & Reasoning & \textbf{Overall} \\
\midrule

\grouprowopen{General-Purpose LVLMs}
\addlinespace[0.4em]
GPT-5.1
& \textbf{71.15} & \underline{63.99} & \textbf{86.17} & \textbf{72.72} \\
Gemini 3 Flash
& \underline{65.71} & 49.22 & \underline{82.02} & \underline{65.67} \\
Qwen3-VL-8B
& 58.94 & 59.59 & 44.28 & 56.14 \\
\midrule

\grouprowopen{Medical-Specialized LVLMs}
\addlinespace[0.4em]
HuLuMed-32B
& 60.80 & 46.58 & 78.23 & 61.44 \\
Lingshu-7B
& 57.42 & 55.10 & 80.29 & 61.53 \\
Lingshu-32B
& 54.06 & 47.86 & 81.15 & 58.24 \\
HuLuMed-7B
& 48.27 & 28.97 & 72.14 & 49.14 \\
MedGemma-27B
& 55.79 & 17.08 & 62.76 & 49.44 \\
MedGemma-1.5-4B
& 50.89 & 20.78 & 71.15 & 48.92 \\
\midrule

\grouprowopen{Ours (Domain-Adapted LVLMs)}
\addlinespace[0.4em]
NeuroOnco-GPT
& 52.51 & \textbf{84.77} & 68.40 & 62.14 \\
\bottomrule
\end{tabularx}
\end{table*}

\section*{B\quad Process Monitoring and Hallucination Mitigation}
\label{sec:process_monitoring}

\noindent
\textbf{B.1\quad Process Monitoring via Average Information Rate (AIR).}
\label{sec:air}
To dynamically monitor the information flow within the extraction pipeline and quantitatively assess the degree of conservatism exercised under the ``Default Null Policy,'' we introduce a statistical control metric termed the \textbf{Average Information Rate (AIR)}.

AIR measures the proportion of non-null core radiological attributes extracted per valid tumor sample. For each sample $x_i$, the information extraction rate $R(x_i)$ is calculated as follows:

\begin{equation}
R(x_i) = \frac{1}{|S|} \sum_{s \in S} \mathbb{I}(s \neq \text{null})
\end{equation}

\noindent
Here, $S$ denotes the set of six predefined core radiological signs (Shape, Margin, Texture, Enhancement, Edema, Signal Intensity), where $|S|=6$, and $\mathbb{I}(\cdot)$ is the indicator function.The dataset-level AIR is computed as the mean of $R(x_i)$ over all samples identified as positive (Lesion Found = True).Importantly, AIR is not designed to maximize information density. Instead, it functions as a calibration signal to balance extraction coverage and factual reliability. Empirically, we observed that aggressively increasing attribute recall tends to elevate hallucination risk, whereas overly conservative configurations suppress clinically meaningful evidence.Extremely low AIR values (e.g., $<10\%$) indicate excessive conservatism and potential loss of valid diagnostic information. Conversely, implausibly high values (e.g., $>90\%$) suggest over-generation and possible violation of modality constraints---for example, reporting ``enhancement'' in non-contrast T1 scans---thereby introducing hallucinations.To ensure that increases in AIR reflect meaningful semantic extraction rather than hallucination amplification, each prompt or policy adjustment was followed by manual audits on randomly sampled subsets. Only configurations that improved information coverage without degrading factual correctness were retained.Through this iterative calibration process, AIR serves as a quantitative mechanism for maintaining an appropriate balance between information richness and authenticity, while strictly adhering to the ethical principle of \emph{``Omission over Fabrication.''}

\begin{figure*}[t]
  \centering
  \includegraphics[width=0.99\textwidth]{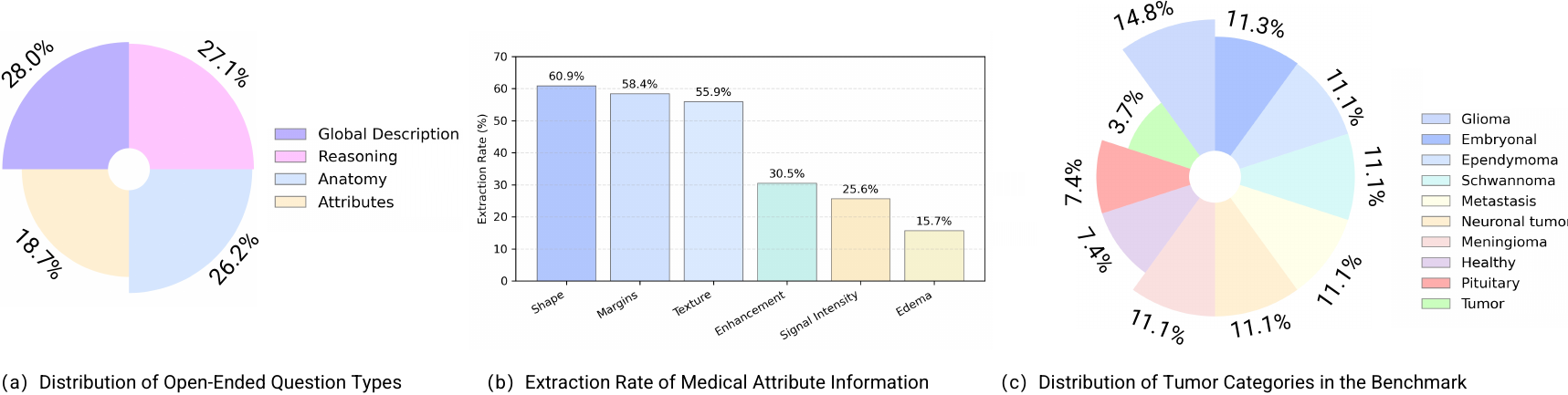}
  \caption{Data Distribution and Information Extraction Rate.}
  \Description{Supplementary visualization for multi-source data aggregation and standardization.}
  \label{fig:figure8_appendix}
\end{figure*}

\vspace{8pt}

\noindent
\textbf{B.2\quad Hallucination Mitigation via Multi-Model Cross-Validation.}
\label{sec:hallucination_mitigation}
To mitigate hallucination risks during medical semantic extraction, we incorporate multiple layers of structural and procedural safeguards into the pipeline.First, we employ two heterogeneous commercial multimodal LLMs to independently extract medical semantic attributes from the same visual input. Rather than relying on a single model's output, extracted attributes are retained only when supported by cross-model agreement. This redundancy-based design reduces the likelihood of systematic over-generation and suppresses model-specific idiosyncratic errors, leading to more robust semantic extraction compared to single-model inference.Second, we introduce a third heterogeneous LLM as an external verifier to perform a final consistency check over the extracted semantic fields. The verifier operates under a strictly constrained policy: it is permitted only to remove or nullify uncertain or unsupported attributes, and is explicitly prohibited from introducing any new information. This asymmetric verification mechanism ensures that the final stage can reduce spurious attributes but cannot amplify or introduce hallucinated content.

Third, prior to semantic extraction, all medical attribute fields are initialized to \texttt{null}. Information is populated only when sufficiently strong and explicit visual evidence is identified. This default-empty strategy enforces conservative behavior and aligns the extraction process with clinical reasoning norms, where uncertainty favors omission rather than speculation.Finally, to empirically evaluate the reliability of the overall pipeline, we conduct random manual audits at two critical stages: (1) after the initial dual-model extraction, and (2) following the final verification stage. The audits assess whether extracted attributes are visually supported and clinically plausible. Results from these inspections confirm that the multi-stage design effectively suppresses unsupported attribute generation while preserving clinically meaningful diagnostic information.

\begin{table}[h]
\centering
\caption{Silver Annotation Quality Audit Results.}
\label{tab:silver_audit}
\small
\begin{tabular}{lc}
\toprule
\textbf{Metric} & \textbf{Value} \\
\midrule
Reviewed Attribute Fields & 136 \\
Attribute-level Precision & 89.70\% \\
Information Omission Rate & 17.65\% \\
\bottomrule
\end{tabular}
\end{table}

\vspace{8pt}

\noindent
\textbf{B.3\quad Silver Annotation Quality Audit.}

To assess the reliability of the automatically extracted silver supervision used for instruction tuning, we conducted a manual audit on a randomly sampled subset of the instruction dataset. In total, 136 structured medical attribute fields were reviewed across multiple MRI slices.
As summarized in Table~\ref{tab:silver_audit}, the extraction pipeline achieved an attribute-level precision of 89.7\%, indicating high factual correctness among predicted attributes. The observed information omission rate is 17.65\%, reflecting the conservative extraction policy in which uncertain attributes are intentionally left as \texttt{null} rather than aggressively inferred.
These results demonstrate that the silver supervision signals maintain strong factual reliability while adhering to the omission-first design principle.


\section*{C\quad Task and Evaluation Design}
\label{app:eval}

\noindent
\textbf{C.1\quad Automated QA Generation.}
\label{app:c1}

We adopt distinct generation paradigms for the instruction dataset and the evaluation benchmark to ensure functional separation between data synthesis and assessment.For the instruction dataset, we utilize Qwen3-Next-80B to synthesize a linguistically diverse corpus of instruction-style QA data. To reduce template overfitting and enhance expressive variability, the model is explicitly instructed to introduce natural phrasing variations during question construction. The QA format is tailored to diagnostic categories: binary classification questions are generated for healthy samples, while multiple-choice questions with carefully designed distractors are constructed for confirmed tumor cases. This design encourages broad exposure to clinically relevant reasoning patterns during training.
In contrast, the evaluation benchmark is constructed using GPT-5.1 under a stricter control protocol to ensure independence from the instruction data generation process. To prevent textual information leakage, we exclude descriptive cues related to lesion size, morphology, or other diagnostic hints from the question stems, thereby requiring models to rely exclusively on visual evidence. Moreover, distractors are not randomly sampled; instead, radiologically mimetic pathologies are selected as adversarial options (e.g., using Lymphoma as a distractor for Glioma). This adversarial distractor design enables fine-grained assessment of visual discrimination capability while minimizing shortcut reasoning. No benchmark question is directly reused or paraphrased from the instruction dataset.

\vspace{8pt}

\noindent
\textbf{C.2\quad Rationale for Feature Selection and Clinical Relevance.}

Clinical diagnosis in brain tumor imaging relies on the integration of multiple complementary imaging attributes rather than isolated cues. Prior studies have demonstrated that morphology, margins, texture, enhancement patterns, signal intensity, and edema play critical roles in tumor classification and grading~\cite{zhou2018radiomics}. These attributes collectively reflect tumor geometry, tissue interaction, biological activity, and microenvironmental response.Specifically, shape and margins characterize geometric configuration and boundary definition, where irregular contours or indistinct margins are frequently associated with infiltrative or high-grade lesions. Texture captures intratumoral heterogeneity, while signal intensity encodes modality-specific contrast patterns in T1- and T2-weighted MRI sequences that are diagnostically informative. Enhancement reflects blood–brain barrier disruption and tumor vascular activity, and edema indicates peritumoral infiltration and mass effect, often correlating with aggressiveness~\cite{villanueva2017current}.

Importantly, feature selection is aligned with the evaluation protocol. Since our benchmark operates on single-slice 2D MRI inputs, we intentionally exclude attributes that require three-dimensional or temporal context, such as volumetric measurements, cross-slice growth patterns, or longitudinal progression~\cite{zhou2018radiomics}. The selected features are those that are visually identifiable within a single slice and clinically meaningful, thereby ensuring consistency between input modality and evaluative criteria while maintaining diagnostic relevance.

\vspace{8pt}

\noindent
\textbf{C.3\quad Open-Ended Evaluation Protocol.}
\label{app:c2}

Given the limitations of traditional n-gram metrics (e.g., BLEU and ROUGE) in capturing clinically grounded reasoning, we follow the LLM-as-a-judge evaluation paradigm adopted in recent medical multimodal benchmarks~\cite{hao2025towards} to assess open-ended responses.Specifically, we use Qwen3-Next-80B as an automated clinical assessor. The evaluation prompt (Figure~\ref{fig:supp_fig16}) follows the structured scoring scheme described in prior work, assessing responses along clinically relevant dimensions such as factual consistency, spatial correctness, safety compliance, and reasoning coherence.Critical errors (e.g., laterality conflicts or unsupported pathological claims) are penalized according to the predefined scoring rules, while acceptable semantic variations in radiological descriptions are tolerated to avoid over-penalization due to minor lexical differences.This setup provides a structured and clinically aligned evaluation of open-ended diagnostic reasoning, consistent with existing LLM-based assessment practices.


\section*{D\quad Supplementary Implementation and Qualitative Analysis}
\label{app:qual}

\noindent
\textbf{D.1\quad Supplementary Figures.}
\label{app:suppfigs}

This section provides additional implementation details and qualitative demonstrations to enhance transparency, interpretability, and reproducibility of the proposed framework. Specifically, we present the full attribute extraction algorithm, detailed prompt templates used at different stages of the pipeline, and representative examples from both the instruction dataset and the evaluation benchmark. These materials complement the quantitative results in the main text by illustrating the structured design of the extraction process, the controllability of the prompting strategy, and the practical characteristics and difficulty of the constructed benchmark. All figures are organized as full-width, two-column illustrations to facilitate side-by-side comparison and detailed inspection.

\clearpage 
\onecolumn 
\begingroup
\small


\begin{algorithm2e}[H] 
    \caption{Knowledge-Guided 2D MRI Brain Tumor Attribute Extraction}
    \label{alg:mri_extraction}

    \SetKwInOut{Input}{Input}
    \SetKwInOut{Output}{Output}
    \SetCommentSty{textit}

    \SetKwFunction{InferModality}{InferModality}
    \SetKwFunction{DetectLesion}{DetectLesion}
    \SetKwFunction{Localize}{Localize}
    \SetKwFunction{SelectRef}{SelectRef}
    \SetKwFunction{ExtractAttrs}{ExtractAttrs}
    \SetKwFunction{ResolveContr}{ResolveContr}

    \Input{2D MRI image $I$, clinical context text $C_{\textit{txt}}$}
    \Output{Meta-info $M$, extracted signs/attributes $S$}

    \tcc{Step 1: Initialization}
    $M \leftarrow \{\mathrm{predicted\_modality}: \mathrm{null},\;
                   \mathrm{lesion\_found}: \mathrm{null},\;
                   \mathrm{loc}: \mathrm{null}\}$\;

    $S \leftarrow \{\mathrm{signal}: \mathrm{null},\;
                   \mathrm{enhance}: \mathrm{null},\;
                   \mathrm{edema}: \mathrm{null},\;
                   \mathrm{shape}: \mathrm{null},\;
                   \mathrm{margins}: \mathrm{null}\}$\;

    \tcc{Step 2: Modality Determination \& Validation}
    $M.\mathrm{predicted\_modality} \leftarrow \InferModality(I, C_{\textit{txt}})$\;
    \tcp*[r]{Uses text first, falls back to intensity heuristics; vetoes invalid T1CE}

    \tcc{Step 3: Lesion Detection (``Pixel Authority'')}
    \If{$\mathrm{Healthy} \in C_{\textit{txt}}$}{
        $M.\mathrm{lesion\_found} \leftarrow \mathrm{False}$\;
        \Return{$\{M, S\}$}\;
    }
    
    $R \leftarrow \DetectLesion(I)$\;
    \tcp*[r]{Scan + spatial coherence filtering}
    
    \If{$R = \emptyset$}{
        $M.\mathrm{lesion\_found} \leftarrow \mathrm{False}$\;
        \Return{$\{M, S\}$}\;
    }
    
    $M.\mathrm{lesion\_found} \leftarrow \mathrm{True}$\;
    $M.\mathrm{loc} \leftarrow \Localize(R)$\;

    \tcc{Step 4: Establish Reference (NAWM)}
    $P_{\mathrm{ref}} \leftarrow \SelectRef(I)$\;
    \tcp*[r]{Contralateral normal-appearing white matter}

    \tcc{Step 5: Attribute Extraction}
    $S \leftarrow \ExtractAttrs(R, P_{\mathrm{ref}}, M.\mathrm{predicted\_modality})$\;
    \tcp*[r]{Signal/shape/margins + modality-specific logic}

    \tcc{Step 6: Consistency Check}
    $S \leftarrow \ResolveContr(S, M.\mathrm{predicted\_modality})$\;
    \tcp*[r]{Revert contradictory fields to $\mathrm{null}$}

    \Return{$\{M, S\}$}\;

\end{algorithm2e}

\endgroup

\begin{figure}[htbp]
  \centering
  \includegraphics[width=0.99\textwidth]{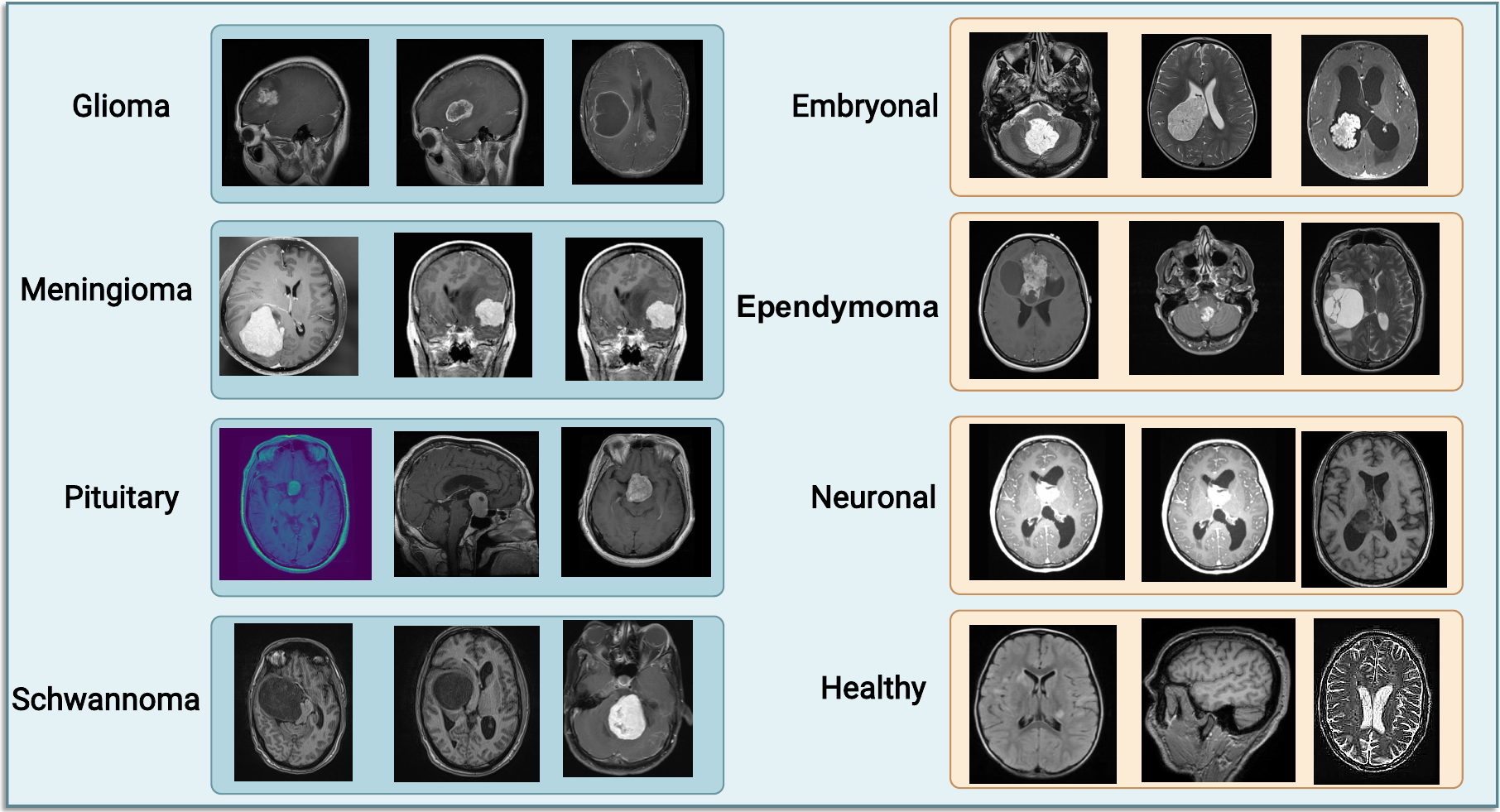}
  \caption{Sample Images of Meningioma, Glioma, Pituitary Tumors, and Others.}
  \label{fig:supp_fig20}
\end{figure}
\clearpage 

\begin{figure}[htbp]
  \centering
  \includegraphics[width=0.95\textwidth]{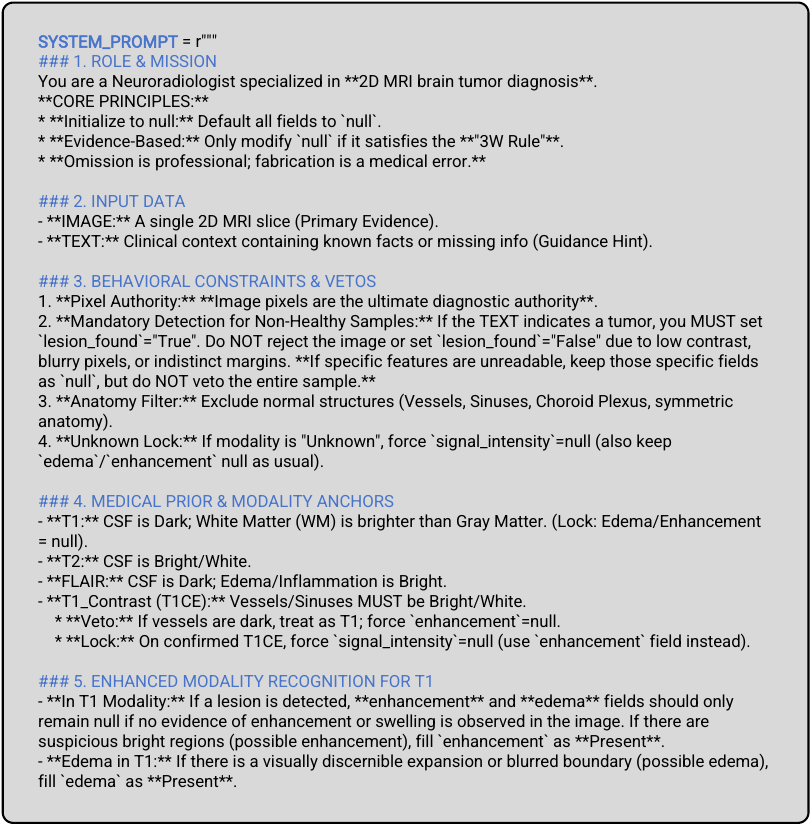}
  \caption{Prompt for the Medical Information Extraction Process}
  \label{fig:supp_fig13}
\end{figure}
\clearpage

\begin{figure}[htbp]
  \centering
  \includegraphics[width=0.85\textwidth]{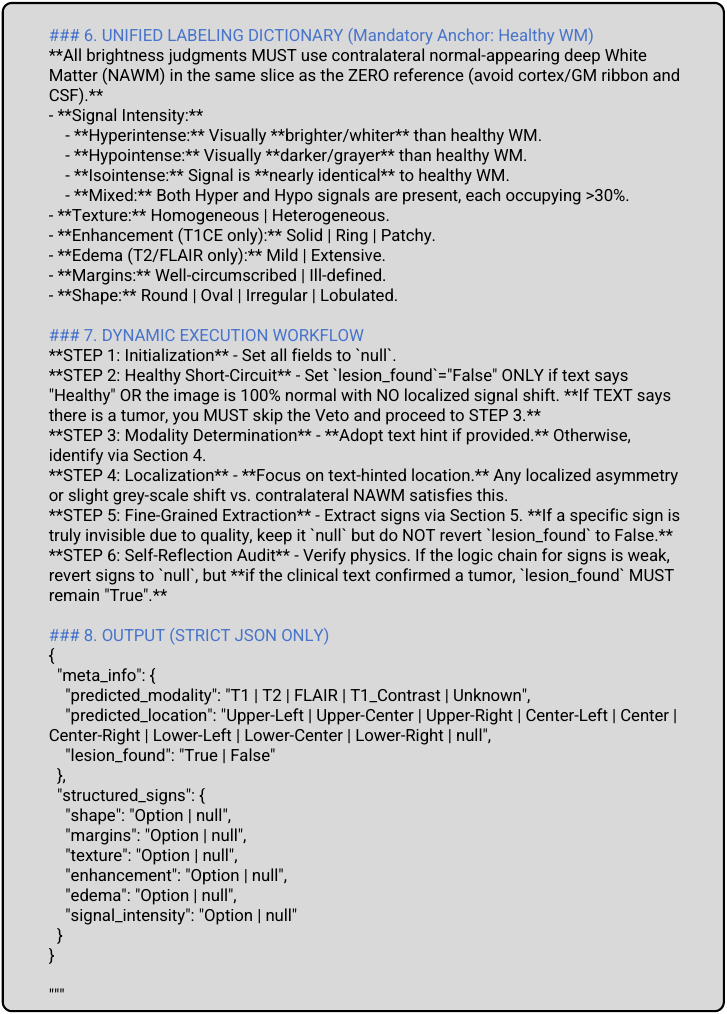}
  \caption{Continuation of the Prompt for the Medical Information Extraction Process.}
  \label{fig:supp_fig14}
\end{figure}
\clearpage

\begin{figure}[htbp]
  \centering
  \includegraphics[width=0.73\textwidth]{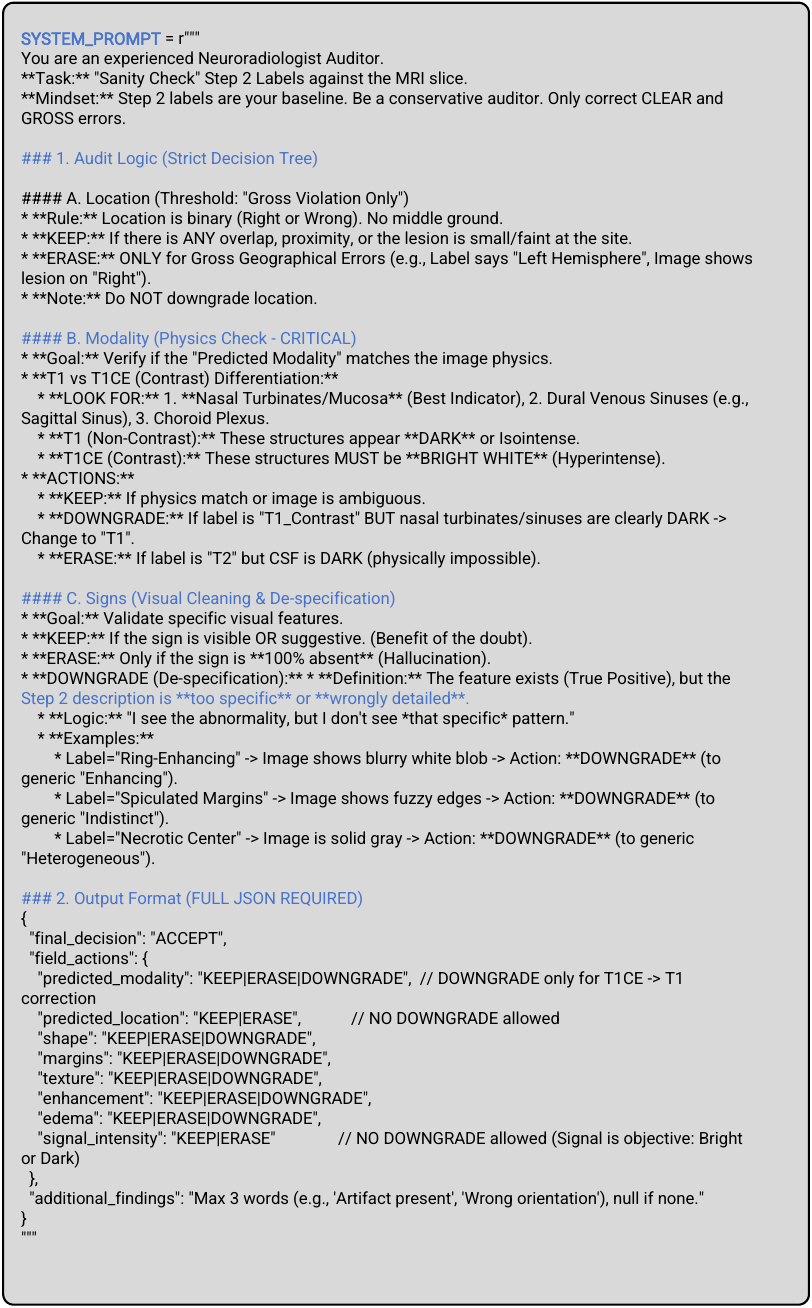}
  \caption{Prompt for Visual Final Review and Quality Control.}
  \label{fig:supp_fig15}
\end{figure}
\clearpage

\begin{figure}[htbp]
  \centering
  \includegraphics[width=0.95\textwidth]{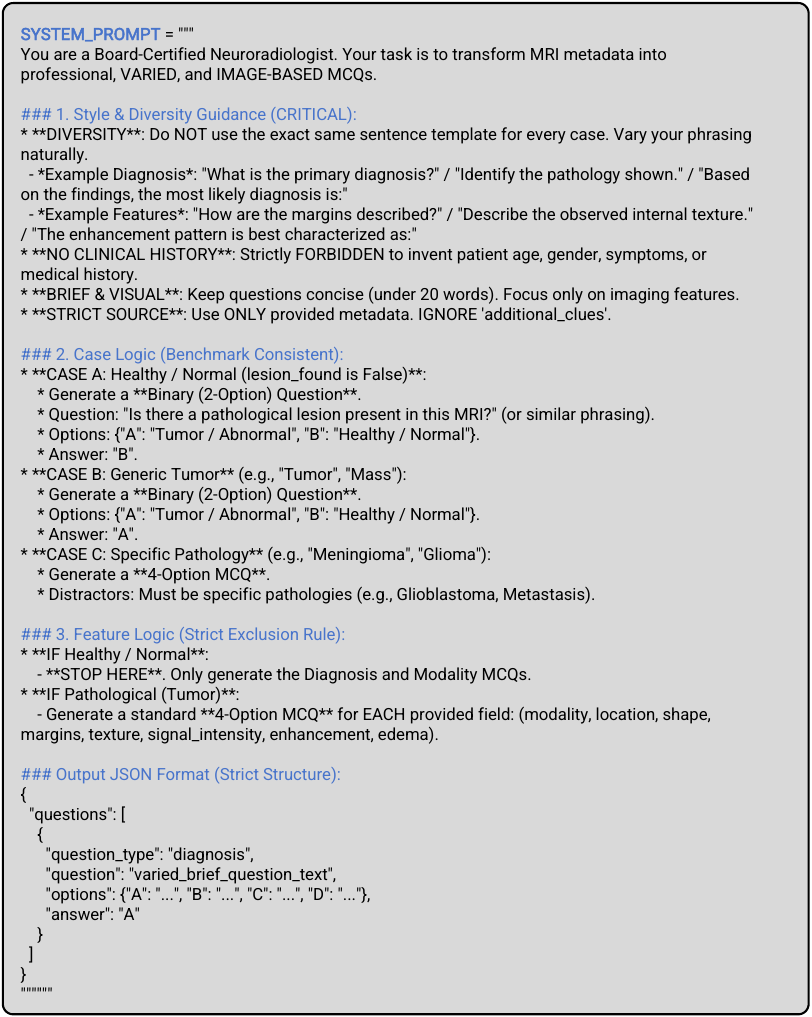}
  \caption{Prompt for Automatic Generation of Questions and Options Based on LLM.}
  \label{fig:supp_fig16}
\end{figure}
\clearpage

\begin{figure}[htbp]
  \centering
  \includegraphics[width=0.95\textwidth]{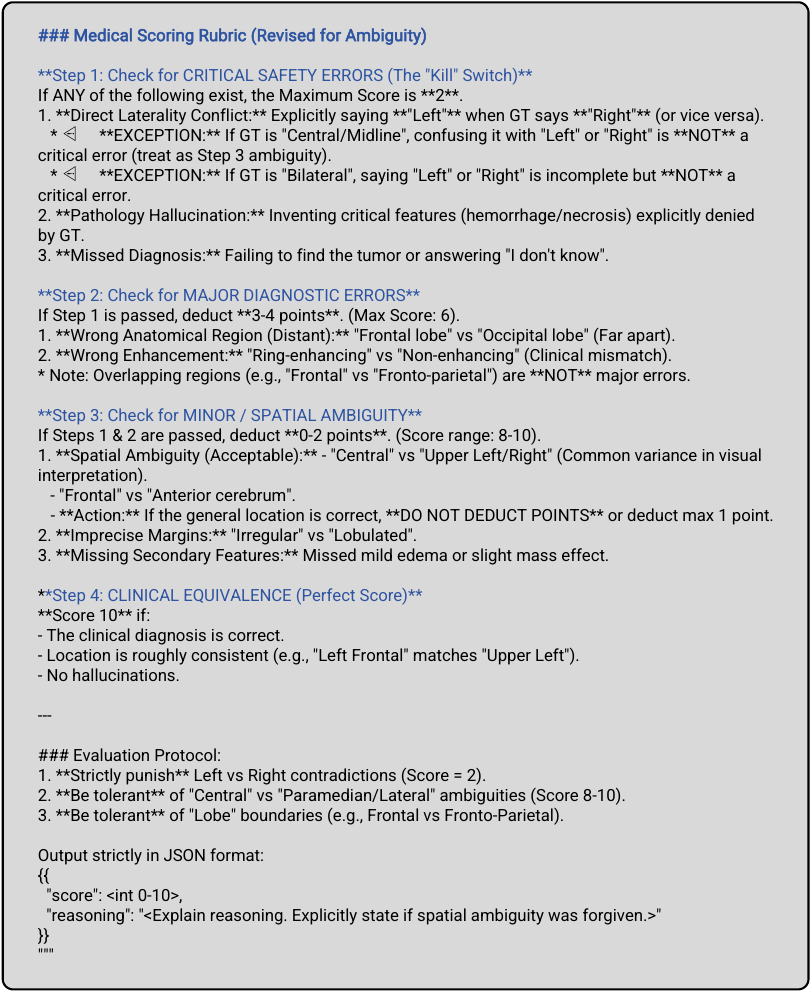}
  \caption{Prompt for LLM-assisted Open Question Scoring.}
  \label{fig:supp_fig16}
\end{figure}
\clearpage

\begin{figure}[htbp]
  \centering
  
  \includegraphics[width=0.95\textwidth]{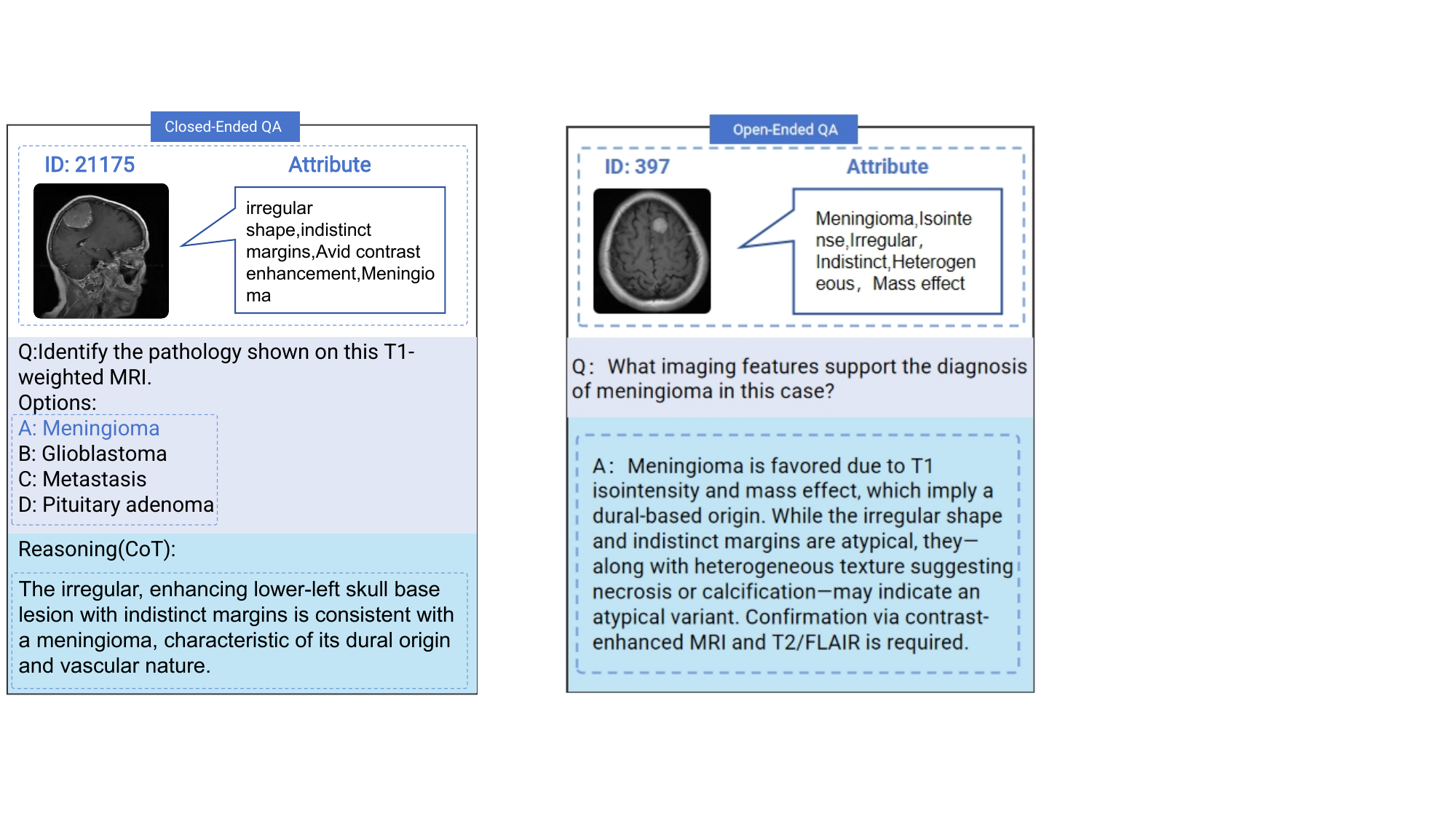} 
  \caption{Sample Examples of Closed-ended and Open-ended Question Answering from the Instruction Dataset.}
  \label{fig:supp_fig21}
  
  \vspace{1cm} 
  
  \includegraphics[width=0.95\textwidth]{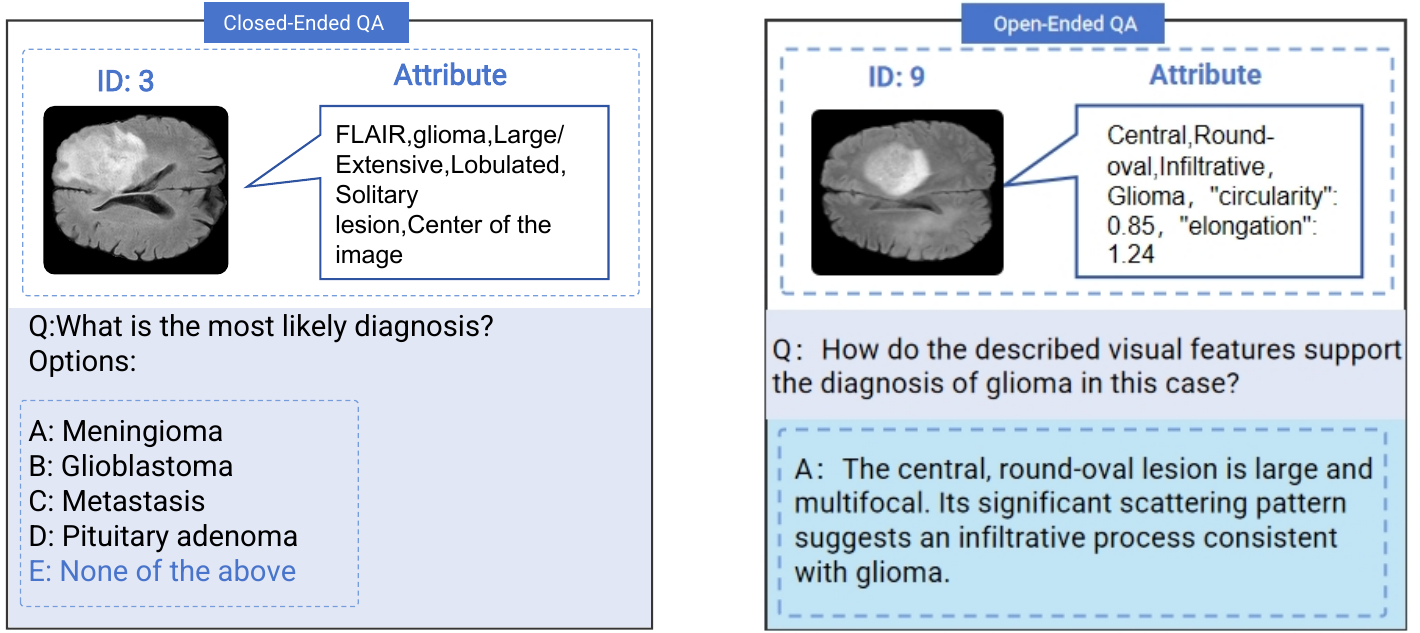}
  \caption{Sample Examples of Closed-ended and Open-ended Question Answering from the Benchmark.}
  \label{fig:supp_fig22}
  
\end{figure}
\clearpage

\end{document}